\documentclass[11pt, letterpaper]{cmu}

\usepackage[all]{hypcap}
\usepackage[numbers,sort&compress]{natbib}
\usepackage{hyperref}[citecolor=magenta]

\hypersetup{
    colorlinks = true,
    citecolor = {magenta},
}

\usepackage{microtype}
\usepackage{graphicx}
\usepackage{subfigure}
\usepackage{booktabs}
\usepackage{float}
\usepackage[font=footnotesize]{caption}
\usepackage{amsmath}
\usepackage{amssymb}
\usepackage{mathtools}
\usepackage{amsthm}
\usepackage{mathrsfs}
\usepackage{nicefrac}
\usepackage{dsfont}
\usepackage{enumitem}
\usepackage{float}

\usepackage{xspace}
\usepackage[capitalize,noabbrev]{cleveref}
\usepackage{subcaption}
\usepackage{wrapfig}
\usepackage{lipsum}
\usepackage{listings}

\usepackage{amsmath}
\usepackage{amssymb}
\usepackage{mathtools}
\usepackage{amsthm}
\usepackage{bbm}
\usepackage{fleqn, tabularx}
\usepackage{multirow}
\usepackage[export]{adjustbox}
\usepackage{colortbl}

\usepackage{algpseudocode}
\usepackage{setspace}

\usepackage{color}
\definecolor{deepblue}{rgb}{0,0,0.5}
\definecolor{deepred}{rgb}{0.6,0,0}
\definecolor{deepgreen}{rgb}{0,0.5,0}
\definecolor{boost_correct_to_correct}{HTML}{66C2A5}
\definecolor{default_correct_to_correct}{HTML}{fc8d62}
\definecolor{dup_correct_to_correct}{HTML}{8da0cb}
\definecolor{new_correct_to_correct}{HTML}{e78ac3}

\newcommand\pythonstyle{\lstset{
basicstyle=\ttfamily\footnotesize,
language=Python,
morekeywords={self, clip, exp, mse_loss, uniform_sample, concatenate, logsumexp},              
keywordstyle=\color{deepblue},
emph={MyClass,__init__},          
emphstyle=\color{deepred},   
stringstyle=\color{deepgreen},
frame=single,                       
showstringspaces=false
}}

\lstnewenvironment{python}[1][]
{
\pythonstyle
\lstset{#1}
}
{}

\newcommand\pythoninline[1]{{\pythonstyle\lstinline!#1!}}

\definecolor{blanchedalmond}{rgb}{1.0, 0.92, 0.8}
\definecolor{carmine}{rgb}{0.59, 0.0, 0.09}
\definecolor{lightblue}{rgb}{0.22,0.45,0.70}% light blue

\renewcommand{\mathbf}{\boldsymbol}

\makeatletter
\def\Ddots{\mathinner{\mkern1mu\raise\p@
\vbox{\kern7\p@\hbox{.}}\mkern2mu
\raise4\p@\hbox{.}\mkern2mu\raise7\p@\hbox{.}\mkern1mu}}
\makeatother
\numberwithin{equation}{section}

\definecolor{amaranth}{rgb}{0.9, 0.17, 0.31}
\definecolor{antiquebrass}{rgb}{0.8, 0.58, 0.46}
\definecolor{antiquefuchsia}{rgb}{0.57, 0.36, 0.51}
\definecolor{chromeyellow}{rgb}{0.31, 0.47, 0.26}

\usepackage[dvipsnames]{xcolor}
\definecolor{maj5}{HTML}{2b8cbe}
\definecolor{maj5Imp}{HTML}{084081}

\definecolor{seq5wo}{HTML}{d95f0e}
\definecolor{seq5woImp}{HTML}{662506}

\definecolor{seq5w}{HTML}{6a51a3}
\definecolor{seq5wImp}{HTML}{3f007d}

\definecolor{selfwo}{HTML}{d95f0e}
\definecolor{selfwoImp}{HTML}{662506}

\definecolor{selfw}{HTML}{6a51a3}
\definecolor{selfwImp}{HTML}{3f007d}

\definecolor{glorewo}{HTML}{d95f0e}
\definecolor{glorewoImp}{HTML}{662506}

\definecolor{glorew}{HTML}{6a51a3}
\definecolor{glorewImp}{HTML}{3f007d}

\definecolor{vstar}{HTML}{d95f0e}
\definecolor{vstarImp}{HTML}{662506}

\makeatletter
\def\mathcolor#1#{\@mathcolor{#1}}
\def\@mathcolor#1#2#3{%
  \protect\leavevmode
  \begingroup
    \color#1{#2}#3%
  \endgroup
}
\makeatother

\usepackage[textsize=tiny]{todonotes}
\usepackage{algorithm}
\usepackage{amssymb}

\usepackage[skins,theorems,breakable]{tcolorbox}
\Crefformat{equation}{#2Eq.\;(#1)#3}

\Crefformat{figure}{#2Figure #1#3}
\Crefformat{assumption}{#2Assumption #1#3}
\Crefname{assumption}{Assumption}{Assumptions}

\usepackage{crossreftools}
\usepackage[suppress]{color-edits}

\usepackage{dsfont}
\usepackage{nicefrac}

\newtcolorbox{analysisbox}[1][]{
    enhanced jigsaw,
    colback=white,
    colframe=blue!75!black,
    fonttitle=\bfseries,
    boxsep=5pt,
    left=5pt,
    right=5pt,
    top=5pt,
    bottom=5pt,
    title=#1,
}

\definecolor{rliableolive}{HTML}{BBCC33}
\definecolor{rliableblue}{HTML}{77AADD}
\definecolor{rliablered}{HTML}{EE8866}

\tcbset{
  aibox/.style={
    width=\linewidth,
    top=8pt,
    bottom=4pt,
    colback=rliableolive!8!white,
    colframe=black,
    colbacktitle=black,
    enhanced,
    center,
    attach boxed title to top left={yshift=-0.1in,xshift=0.15in},
    boxed title style={boxrule=0pt,colframe=white,},
  }
}
\newtcolorbox{AIbox}[2][]{aibox,title=#2,#1}
\definecolor{lightblue}{rgb}{0.22,0.45,0.70}% light blue

\usepackage{pifont}
\usepackage{soul}
\definecolor{highlightmistake}{RGB}{255, 179, 179}
\definecolor{highlightcorrect}{RGB}{179, 255, 179}

\usepackage{hyperref}[citecolor=magenta,linkcolor=magenta]
\usepackage{enumerate}
\usepackage{enumitem}
\hypersetup{
    colorlinks = true,
    citecolor = {magenta},
}
\usepackage{bbm}

\usepackage{arydshln}
\usepackage{longtable}
\usepackage{xcolor,colortbl}
\usepackage{listings}
\newcommand{\citeg}[1]{\citep[e.g.,][]{#1}}
\usepackage{amsthm}
\usepackage{algorithm} % For the algorithm environment
\usepackage{algpseudocode} % For the algorithmic environment

\newcommand{\methodname}{\emph{\textbf{TTI}}}

\definecolor{lightblue}{rgb}{0.22,0.50,0.70}

\DeclareMathOperator*{\argmax}{arg\,max}

\title{Thinking vs. Doing: Agents that Reason by  Scaling Test-Time Interaction}

\author[1,2]{Junhong Shen*}
\author[3]{Hao Bai*}
\author[4]{Lunjun Zhang}
\author[5]{Yifei Zhou}
\author[1]{Amrith Setlur}
\author[7]{Shengbang Tong}
\author[6]{Diego Caples}
\author[3]{Nan Jiang}
\author[3]{Tong Zhang}
\author[1]{Ameet Talwalkar}
\author[1]{Aviral Kumar}

\affil[1]{Carnegie Mellon University}
\affil[2]{Scribe}
\affil[3]{University of Illinois Urbana-Champaign}
\affil[4]{University of Toronto}
\affil[5]{University of California, Berkeley}
\affil[6]{The AGI Company}
\affil[7]{New York University}

\correspondingauthor{junhongs@andrew.cmu.edu, haob2@illinois.edu}

\vspace{-0.2cm}
\begin{abstract}
\textbf{Abstract:} The current paradigm of test-time scaling relies on generating long reasoning traces (``thinking'' more) before producing a response. In agent problems that require interaction, this can be done by generating thinking traces before acting in the world. However, this process does not allow agents to acquire new information from the environment or adapt their behavior over time. In this work, we propose to scale \textbf{test-time interaction}, an untapped dimension of test-time scaling that increases the agent's interaction horizon to enable running rich behaviors such as exploration, backtracking, and dynamic re-planning within a single rollout. To demonstrate the promise of this scaling dimension, we study the domain of web agents. We first show that even prompting-based interaction scaling without any training can improve task success on web benchmarks non-trivially. Building on this, we introduce \methodname{} (Test-Time Interaction), a curriculum-based online reinforcement learning (RL) approach that trains agents by adaptively adjusting their rollout lengths. Using a Gemma 3 12B model, \methodname{} produces state-of-the-art open-source, open-data web agents on WebVoyager and WebArena benchmarks. We further show that \methodname{} enables agents to balance exploration and exploitation adaptively. Our results establish interaction scaling as a powerful, complementary axis to scaling per-step compute, offering new avenues for training adaptive agents.

\vspace{2mm}
\textbf{Project page:} \href{https://test-time-interaction.github.io}{https://test-time-interaction.github.io}

\textbf{Code:} \href{https://github.com/test-time-interaction/TTI}{https://github.com/test-time-interaction/TTI}
\end{abstract}

\begin{document}

\maketitle

\vspace{-0.25cm}
\section{Introduction}
\vspace{-0.2cm}

Recent advances in foundation models have enabled a shift from static language models to interactive agents that perform multi-step tasks in dynamic environments like browsers~\citep{WebVoyager,zhou2024webarena,claudecomputer,openaioperator,browser_use2024,hong2023cogagent}, terminals~\citep{claudecode}, and the physical world~\citep{Intelligence2025pi_05AV, Zhai2024FineTuningLV, yuan2021emergence, yuanital, mixtureofmamba,Shi2025HiRO}.
These agents operate in closed-loop settings where each action changes the current state of the world and affects future interaction with the environment. 
As a result, interactive agents must plan under uncertainty and adapt to failures in real time to be successful. How can we build agents that succeed in such interactive settings?

\looseness=-1
Current post-training approaches produce \emph{reactive} agents that respond to immediate observations but struggle with evolving or uncertain task dynamics. Methods like supervised fine-tuning (SFT) on expert demonstrations~\citep{Shen2024ScribeAgentTS, deng2023mindweb, shen2023cross, llmtags, Zhang2025SymbioticCF} or  reinforcement learning (RL) with task rewards~\citep{Zhou2024PAE, Bai2024DigiRLTI, Bai2025DigiQLQ, Shen_Yang_2021,qi2025webrltrainingllmweb} typically train agents to predict a \textit{single best action} at each step. Even with test-time scaling, where agents are prompted to ``think'' longer before prescribing an action~\citep{claudedeepresearch, openaideepresearch, geminideepresearch}, they are still optimized to select the most effective action based on the agent's internal state. While sufficient  for fully observable and stationary tasks, {reactive} policies based on the agent's internal estimate of the task state  are often suboptimal in partially observable (e.g., incomplete details visible on a page) or non-stationary  (e.g., fluctuating prices during flight booking) settings, where adaptive, information-seeking behavior is critical.

In this paper, we argue that instead of reactive ``optimal'' policies, agents should learn adaptive policies that can collect  new information from the environment and adjust their behaviors on-the-fly. A pre-requisite for such adaptability is the ability to \textit{take more actions} during deployment than those prescribed by an expert trajectory. We therefore propose a new dimension of test-time scaling: \textit{\textbf{increasing the number of interaction steps of the agent}}. This allows agents to have sufficient context and time to attempt different behaviors. For example, in a hotel booking task, an agent must first browse many listings to compare user reviews and check availability before selecting the best option. Interaction scaling is orthogonal to existing methods based on chain-of-thought (CoT), which emphasize deeper reasoning per step but do not support information-gathering from the environment. This notion of information gain is unique to agentic tasks with partial observability and requires interaction, not merely larger per-step compute. For instance, an agent that reasons deeply about one selected hotel without interacting further may miss better options that show up only after  exploration.

Although the idea of interaction scaling is conceptually straightforward, extending it to post-training and teaching  agents to scale interaction autonomously
presents key challenges.  
Without appropriate training signals, agents may overfit to exploratory behaviors like blindly clicking links but not making progress toward the actual task objective, wasting the additional steps. To tackle this issue, we propose to combine online RL with a curriculum that prescribes how to scale the interaction horizon, training agents that first learn effective exploitation before extending their horizon to explore.

We instantiate our approach in the domain of web agents, a widely applicable setting with well-established benchmarks. We first show that scaling test-time interaction via prompting the agent to ``think and act again'' after it decides to terminate can already improve the task success rate from 23\% to $\geq$ 28\% on WebArena~\citep{zhou2024webarena} (see Figure~\ref{fig:scaling} for details). While this increases trajectory length and the number of tokens generated, spending an equivalent amount of compute on conventional test-time scaling methods like forcing the agent to think for longer~\citep{Muennighoff2025s1ST} or running best-of-$n$~\citep{wu2025inference, Chow2024InferenceAwareFF, snell2025scaling} yields less than a 3\% gain.  These findings validate interaction scaling as a promising and complementary axis of  test-time scaling.

We then move beyond prompting and develop \methodname{} (Test-Time Interaction), 
a curriculum-based RL approach that trains agents to adaptively scale interaction by gradually increasing the rollout horizon. 
We  scale \methodname~to large and  diverse training sets ($>$100K tasks across $\sim$20 domains) by integrating it with an automated pipeline that generates synthetic tasks for online data collection. \methodname~achieves {state-of-the-art performance among open-source agents trained on open data} on both WebVoyager~\citep{WebVoyager} and WebArena~\citep{zhou2024webarena}, using only a 12B Gemma 3 model, improving over the non-fine-tuned agent by {9\%} and 8\%, respectively.
Our analysis further shows that curriculum training enables adaptive exploration: agents learn to initiate new searches or backtrack in complex tasks, while following efficient paths in simpler ones.

\textbf{In summary}, we introduce interaction scaling as a new dimension for test-time scaling of agents. We propose \methodname{} to train agents by adjusting interaction horizon dynamically.  Our results show that \methodname{} yields strong empirical gains and offers promising directions for domains beyond web navigation.
\vspace{-0.2cm}
\section{Related Work}
\vspace{-0.2cm}
\textbf{Scaffolded foundation models as web agents.} 
Prior works use external control structures to scaffold foundation models via modular prompting~\citep{liu2023webglm,hong2023cogagent, webgum, AgentOccam, zhang2024webpilot, shen2024ups,shen2025cat,Xu2024AgentTrekAT}, programs~\citep{guiapi, xu2024specialized, shenrecon, li2025codepdeinferenceframeworkllmdriven,Zheng2025SkillWeaverWA}, or feedback mechanisms~\citep{pan2024autonomous,shen2022dash, nasbench360,fu2024autoguide}. These methods often rely on proprietary models like GPT-4~\citep{gpt4} or Claude~\citep{claude3}. Thus, progress is driven by designing better prompts and workflows for planning~\citep{sodhi2024step, Abuelsaad2024AgentEFA, l2g,erdogan2025planandact}, self-correction~\citep{Kumar2024TrainingLM}, self-evaluation~\citep{koh2024tree, zheng2024synapse}, or by integrating external modules such as memory~\citep{awm2024wang} or retrieval systems~\citep{Reddy2024InfogentAA}. More recently, developing specialized agents has become a promising direction. ScribeAgent~\citep{Shen2024ScribeAgentTS} first uses real-world data to demonstrate that simple fine-tuning can outperform most prompt-based agents. Prior work also builds automated data curation workflows~\citep{murty2024bagel, NNetscape, Trabucco2025TowardsIT} and distillation methods~\citep{Zhang2025SymbioticCF}. Despite these efforts, scaffolding methods remain fundamentally limited: they do not enable agents to self-improve through interaction, and rely on fixed wrappers that lack adaptability across diverse environments.

\textbf{RL training for foundation model agents.} RL-based approaches enable agents to autonomously improve through interaction. Prior work has explored DPO~\citep{Putta2024AgentQA}, actor-critic~\citep{Bai2024DigiRLTI,Bai2025DigiQLQ, Zhou2024ArCHerTL}, or distributed sampling~\citep{Wang2024DistRLAA}. Pipelines like PAE~\citep{Zhou2024PAE} and Learn-By-Interact~\citep{Su2025LearnbyinteractAD} support automatic task generation, exploration, and labeling. However, most of these approaches lack mechanisms for test-time exploration, limiting the agent's ability to adapt its behavior over long horizons,  especially under partially observable conditions. As \citet{Bai2024DigiRLTI} note, continued training after deployment is often required just to maintain performance with these methods.
Our work addresses this limitation by scaling test-time interaction as an independent dimension, allowing agents to refine behavior while acting. Curriculum-based RL has been applied in AutoWebGLM~\citep{lai2024autowebglm} and WebRL~\citep{qi2025webrltrainingllmweb}, where curricula are based on estimated task difficulty  derived from the complexity of LLM-generated instructions. While our approach also induces a progression from simpler to more complex behaviors, it does so by gradually increasing the allowed interaction horizon rather than relying on explicit measures of task difficulty.

\textbf{Scaling test-time compute. } 
Increasing test-time compute via best-of-$n$ sampling~\citep{Chow2024InferenceAwareFF}, beam search~\citep{Snell2024ScalingLT, Wu2024InferenceSL}, or verifiers~\citep{Cobbe2021TrainingVT, Setlur2024RewardingPS,Setlur2025ScalingTC} has shown to improve  performance in reasoning-heavy tasks. In non-interactive settings like math and competitive coding, recent methods train models   to generate long CoT and scale reasoning internally~\citeg{DeepSeekAI2025DeepSeekR1IR, Muennighoff2025s1ST, Qu2025OptimizingTC, wang2025ragen}. As for multi-turn interactive settings, most existing works  simply integrate CoT prompting into the agent system to enhance per-step reasoning~\citeg{react, erdogan2025planandact}. EXACT \citep{Yu2024ExACTTA} scales up the search process for each action, GenRM-CoT \citep{genrm} the number of verifiers, and \citet{Jin2025TwoHA} the number of agents. However, none of these efforts studies the benefits of scaling over the time horizon, where the agent can explore alternatives, backtrack, or gather more information before committing to certain actions. 
Our work extends this line of research by introducing test-time scaling of interaction. As we will show in our empirical results (Section~\ref{sec:paradign:compare}), the benefits of scaling test-time interaction go beyond test-time scaling (or ``reasoning'') before taking an action, within a given time step, because each extra step of interaction with the environment provides new information to the agentic policy, whereas thinking for longer simply reorganizes information that the agent already has.
\vspace{-0.3cm}
\section{Problem Setup}
\vspace{-0.2cm}
\newcommand{\horizon}{h}
We consider solving a web task as a finite-horizon sequential decision-making process guided by a reward objective\footnote{While this work centers on web agents, we believe the insights should generalize to other agent problem domains, and we hope future work will extend these ideas beyond web agents and web navigation.}. Formally, the environment implements a transition function that   evolves over time and provides an observation $o_t$ at   step $t$ reflecting the current task state (details regarding the parameterization of the observation space are shown below). 
The agent policy $\pi$ is parameterized by a multi-modal foundation model that maps observation history $o_{1:t-1}$ and action history $a_{1:t-1}$ to the next action $a_t$. These histories allow the policy to represent rich, context-dependent behaviors enabled via interaction (details about the design of the action space are shown below). 
We denote the environment horizon, or the maximum number of interaction steps allowed in the environment, as $\horizon$. 
For each task, the actual interaction process ends when the agent issues a stop signal  or reaches the step limit $\horizon$.  Let $h_\mathrm{stop} \in (0, \horizon]$ denote the actual number of steps taken. The agent receives a reward of 1 for task success, and 0 otherwise.

\textbf{Observation space design.} Following~\citep{Shen2024ScribeAgentTS,Zhou2024PAE,zhou2024webarena}, we consider an observation  space consisting of the task goal, the   URL, and a structured representation of the current web page that includes both the accessibility  tree of the web page and a screenshot augmented with a set-of-marks overlay~\citep{WebVoyager}, which assigns a unique identifier to each element that the agent can interact with.  While  the agent has access to all past observations in principle, doing so quickly exhausts the context window in practice, so we truncate observation history to the most recent three steps, similar to prior works~\citep{Zhou2024PAE}. However, the agent still has access to all of its past actions. We show in Appendix~\ref{app:sec:observation}
 an example observation.

\textbf{Action space parameterization.} We adopt a discrete action space with six actions: click, type, scroll, go back, search (e.g., Google or Bing), and stop the task with an answer. Following~\citet{AgentOccam}, we do not consider compound actions like \texttt{goto[url]}, as  complex action spaces can hinder performance. For a detailed definition of each action, see the agent prompt in Appendix~\ref{app:prelim:prompt}.

\vspace{-0.3cm}
\section{Scaling Test-Time Interaction: A New Dimension of Agent Scaling}
\label{sec:paradigm}
\vspace{-0.2cm}
\looseness=-1
Prior methods for LLM test-time compute scaling usually scale the number of thinking tokens at each step~\citep{Shao2024DeepSeekMathPT, chen2025research, wan2025thinktwiceactonce, wang2025ragen}, but this does not enable an agent to engage in longer interactions with the environment to collect new information. In principle, scaling the maximum number of interaction steps should allow the agent to employ richer behavioral strategies such as re-attempts, backtracking, and recovery. We will now verify this hypothesis via controlled experiments on WebArena~\citep{zhou2024webarena}. We will then build upon these insights to develop \methodname, an online RL method to explicitly train agents to optimize test-time interactions.

\looseness=-1
\textbf{Controlled experiment setup.} We choose WebArena~\citep{zhou2024webarena} as our testbed, primarily because it enables reproducible interaction with diverse domains (OneStopShop, Reddit, GitLab,  CMS, and OpenStreetMap) and is equipped with ground truth evaluators. We randomly sample 62 tasks for testing and reserve the remaining 750 for online training (see Section~\ref{sec:tti:challenges}). To ensure sufficient interaction, we set a generous test-time limit of $\horizon = 30$, which is well above the average length of around 6 steps required by most tasks~\citep{awm2024wang,Wang2025InducingPS}. Note that experiments in this section are for analysis rather than benchmarking purposes. We will show more experimental results on multiple benchmarks in Section~\ref{sec:exp}.

\begin{wraptable}{r}{0.2\textwidth}
\vspace{-0.4cm}
\caption{\footnotesize{\textit{\textbf{Base results averaged over three runs.}}}}
\vspace{-0.2cm}
\label{table:prompt}
\Huge
  \centering
\resizebox{0.99\linewidth}{!}{\begin{tabular}{lcc}
\toprule
{\bf Prompt}   &{\bf Task SR (\%)}
\\ \midrule
Action Only   & 14.76\\
 CoT   & 23.81 \\
\bottomrule
\end{tabular}}
\vspace{-0.5cm}
\end{wraptable}
To study the effect of increasing $h$, we use a simple prompting-based agent with Gemma 3 12B~\citep{Kamath2025Gemma3T} base model, which observes the web page and outputs an action via a \textit{single} model call. It does not leverage any retrieval, verifiers, or other external modules, ensuring   any performance gains come solely from increased $h$ but not auxiliary scaffolding.  We also study whether it is beneficial to prompt the agent to generate a reasoning trace before acting  (see Appendix~\ref{app:prelim:prompt} for the templates). As Table~\ref{table:prompt} shows, CoT prompting yields significantly higher task  success rate (SR) than direct action generation, setting a baseline of 23.81\% on the   test tasks. While we also tried  more complex prompts that explicitly ask the agent to summarize the state, assess  progress, and   reflect, they did not yield significant gains (see Appendix~\ref{app:prelim:cot}). We thus adopt a simple chain-of-thought (CoT) approach as the default prompting strategy for experiments in this section.

\vspace{-0.2cm}
\subsection{Scaling Test-Time Interaction by Acting Longer}
\vspace{-0.2cm}

\begin{wrapfigure}{r}{0.55\textwidth}
\vspace{-0.7cm}
\centering
\includegraphics[width=0.99\linewidth]{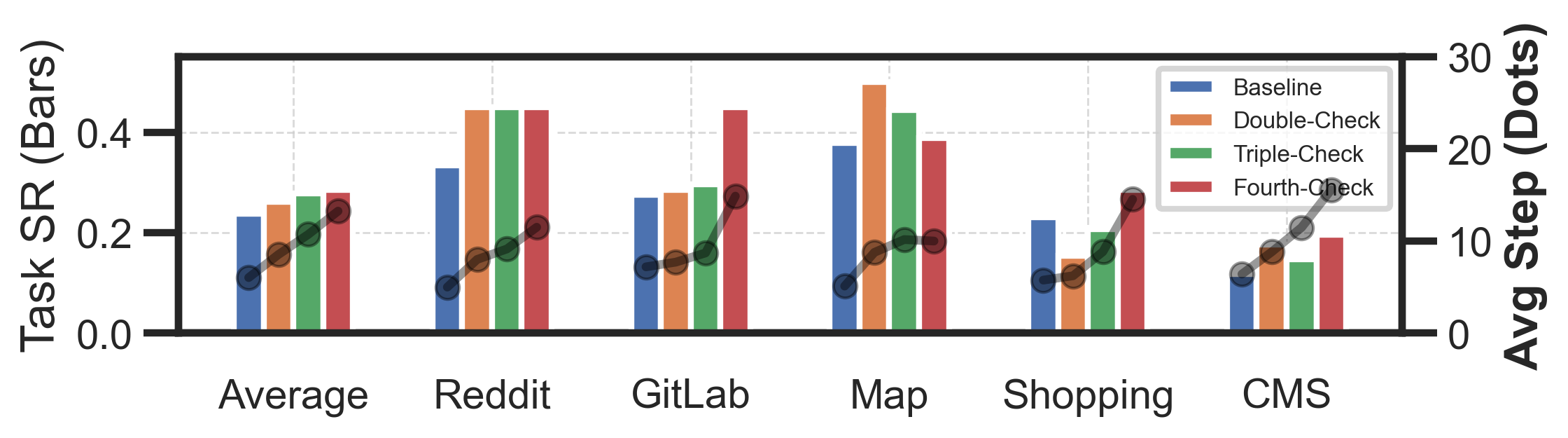}
\vspace{-0.25cm}
    \caption{\footnotesize{\emph{\textbf{Scaling test-time interaction by prompting the agent to ``re-check'' its answer.}} More re-checks lead to longer   trajectories (dots) and higher task success rates (bars), indicating \emph{at least} some correlation between acting for longer and higher success rates.}}    \label{fig:doublecheck}
    \vspace{-0.35cm}
\end{wrapfigure}
To study the impact of test-time interaction scaling, we introduce a purely inference-time ``check-again'' mechanism: after the agent issues the task completion action, we explicitly prompt it to reconsider its decision   by
``\textit{You just signaled task completion. Let's pause and think again...}''
We can extend re-checking from double-check (two passes) to triple-check (three passes) and beyond, using slightly varied prompt phrasings for each pass. Detailed prompts are in Appendix~\ref{app:prelim:scaling}.

As shown in Figure~\ref{fig:doublecheck}, prompting the agent to re-check not only increases the actual interaction length $\horizon_{stop}$ (line plots), 
but also  improves the   success rates on most WebArena domains (bar plots).  When being asked to  ``check-again'', the agent either reaffirms its decision (e.g., \textcolor{ForestGreen}{\small ``\textit{I previously stated the driving time was approximately 30 minutes....30 minutes seems plausible with typical traffic conditions. I'll stick with my previous answer.}''}) or revises it upon reflection (e.g., \textcolor{ForestGreen}{\small ``\textit{My apologies. I jumped to a conclusion prematurely. Although the address book *displays* the desired address, the task requires me to *change* it....I should click button [24] to modify the address.}''}).
In particular, it changes its action $\sim$25\% of the time after double-checking. \textbf{This highlights the potential of interaction scaling:} when given sufficient time, the agent is  likely to explore alternatives before reaching an answer. The chances of the answer being correct could thus be higher.

However, we do observe that repeatedly prompting the agent to re-check can sometimes lead to confusion or hallucination, causing it to revise correct answers into incorrect ones. This may explain the performance drop observed in domains like Map. This is perhaps an inevitable limitation of scaling test-time interaction via \emph{prompting} alone, akin to how prompting is not an effective way to even scale per-step test-time compute (see self-correction results for prompted models in \citet{qu2024recursive}).  We discuss this limitation further in Section~\ref{sec:paradign:compare} and address by \textit{training} the agents explicitly.

\vspace{-0.2cm}
\subsection{Scaling Test-Time Interaction vs. Per-Step Test-Time Compute}
\label{sec:paradign:compare}
\vspace{-0.2cm}

Next, we examine the effect of scaling interaction compared to scaling per-step reasoning: Given a  total token budget, should agents prioritize more interaction steps or generating longer reasoning traces at each step?
To explore this, we study two conventional test-time compute scaling methods.

\looseness=-1
\textbf{Per-step budget forcing.} Following~\citet{Muennighoff2025s1ST}, we prompt the agent to  ``wait and think again'' after generating an initial CoT, encouraging more intermediate reasoning before it commits to an action.  We vary the number of forced waits  from 1 to 4. Despite trying various prompts to induce longer thinking from the agent (see Appendix~\ref{app:prelim:scaling}), our agent changes its actions only 15\% of the time, and most of these changes involve reordering subtasks rather than exploring alternative paths or error correction.

\textbf{Per-step best-of-$n$.} At each step, we sample $n\in\{3,5,7\}$ candidate actions and select one via majority voting, similar to~\citep{Shen2024ScribeAgentTS}. 
We did not scale $n$ up further because best-of-$n$ is compute-intensive for multi-step settings ($n\cdot{h}$ more expensive than the baseline per rollout), 
and we observe diminishing returns from sampling more actions, despite  tuning sampling hyperparameters such as the temperature.

\begin{wrapfigure}{r}{0.46\textwidth}
\centering
        \vspace{-0.2cm}
\includegraphics[width=0.99\linewidth]{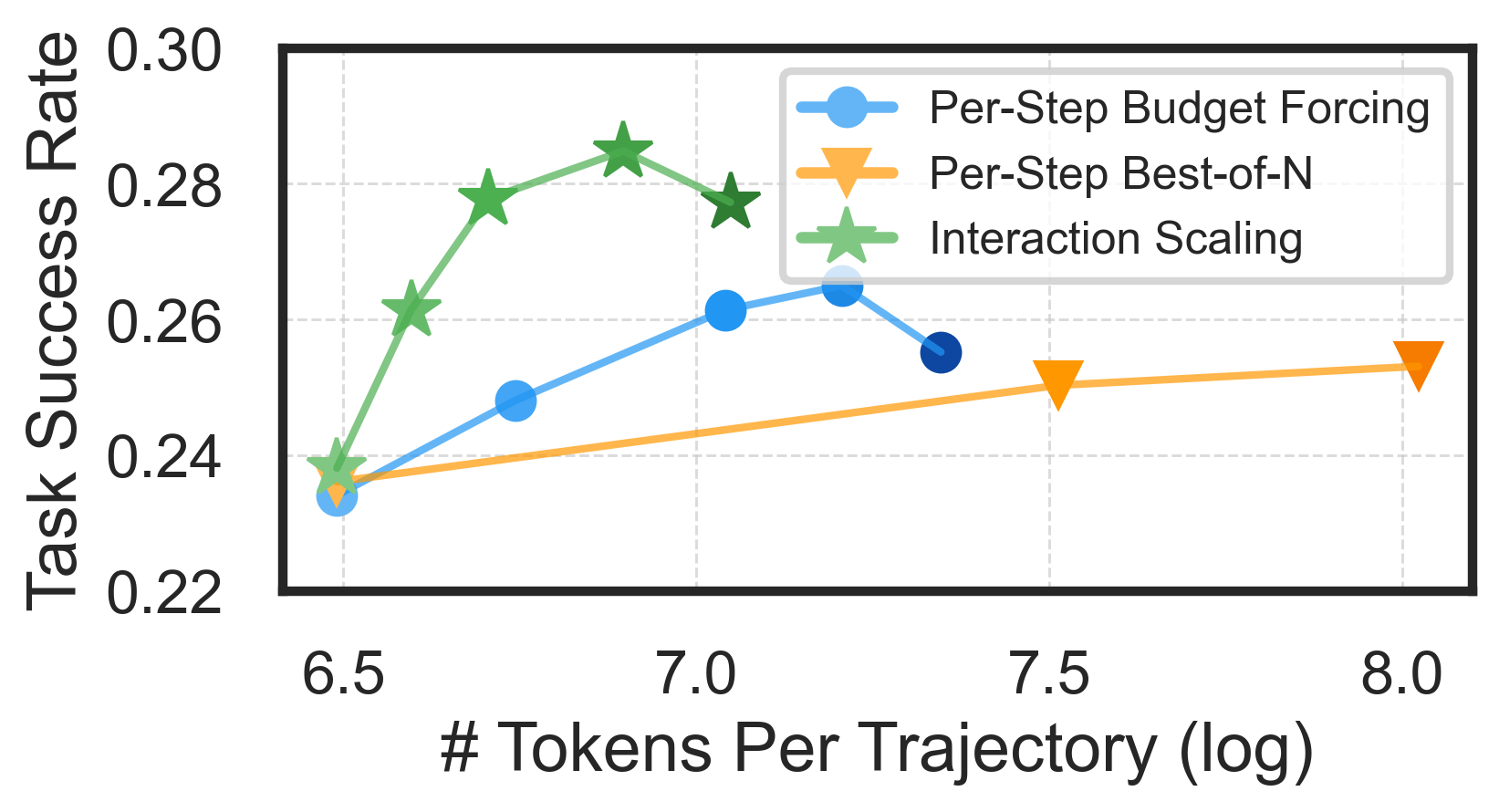}
\includegraphics[width=0.99\linewidth]{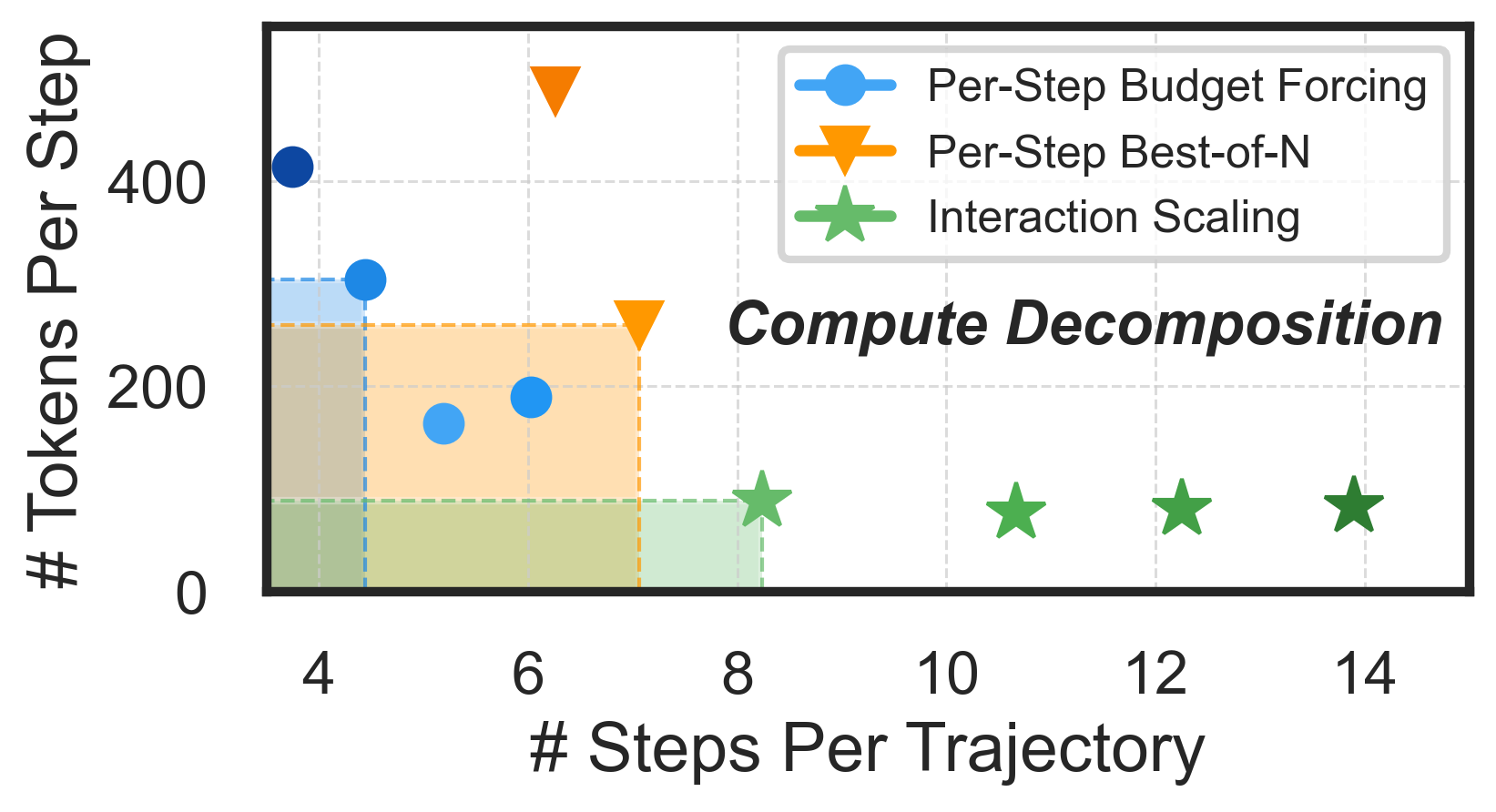}
\vspace{-0.7cm}
    \caption{\footnotesize{
    \emph{\textbf{Top: Task success rate vs. total compute cost}} (measured in tokens per trajectory, log scale). Interaction scaling consistently achieves the highest success rate for a given compute budget.
    \emph{\textbf{Bottom: Decomposition of compute}} into tokens per step and number of interaction steps (excluding the initial point for each approach). Interaction scaling achieves superior performance by distributing compute across more steps with lower per-step cost, in contrast to methods that put more tokens in fewer steps.}}
    \label{fig:scaling}
    \vspace{-0.5cm}
\end{wrapfigure}

\underline{{\textbf{Finding 1: Gaining new information via interaction beats thinking more a single step.}}}  Figure~\ref{fig:scaling} (top) plots the task success   against total compute, measured by the number of tokens per trajectory in log scale. Among the three  strategies,   interaction scaling   (\textcolor{ForestGreen}{green}) shows the steepest upward trend, achieving
the highest success rate as the allowed token budgets increase. Budget forcing (\textcolor{blue}{blue}) yields moderate gains but plateaus around 0.26. Despite incurring the highest cost, best-of-$n$ (\textcolor{orange}{orange})  brings   the least improvements, suggesting that repeatedly sampling actions per step is a less effective use of compute in interactive tasks.

A natural question that arises here is: \emph{how should we distribute a bounded compute budget between running more interaction steps vs. reasoning longer?} These two dimensions present different costs per unit, which may not be known apriori. Hence, we illustrate how scaling the number of interaction steps and the number of tokens per step affect performance and the total number of tokens individually. Figure~\ref{fig:scaling} (bottom) decomposes total compute into tokens per step (y-axis) and steps per rollout (x-axis), so the shaded areas indicate the average total compute required per trajectory. Interaction scaling extends along the x-axis, while per-step reasoning scales along y-axis. We find that scaling \textit{across} steps is more effective than scaling \textit{within} steps in WebArena tasks  (Figure~\ref{fig:scaling}, top), likely because the former enables the agent to gather new information and enrich its context. This ability to query and observe external feedback is unique to agentic settings but not single-turn question-answering tasks. While standard per-step reasoning is constrained by the information already available at each step, our approach takes advantage of this dynamic interaction.  However, in principle, one can combine more reasoning per-step with more interaction, at the cost of spending many tokens.

\underline{{\textbf{Finding 2: Prompting alone is insufficient for interaction scaling.}}} While our results highlight the potential of scaling interaction, the  ``check-again'' strategy   only allows the agent to revisit its behavior upon task completion, it does not enable it to implement nuanced behaviors such as switching between exploration and exploitation in the middle of a rollout. We also experimented with combining interaction scaling with budget forcing and best-of-$n$ (Appendix Table~\ref{app:table:multiplescaling}) and observe that simply increasing test-time compute via prompting does not yield additive gains. In fact, the final downward trend in Figure~\ref{fig:scaling} (top) suggests that asking the agent to re-check too many times or think for too long can confuse it and degrade performance. This shows the need for methods that \textit{train} agents to optimize for best behavior when scaling test-time interaction, rather than na\"ive prompting.

\begin{AIbox}{Takeaways: Scaling test-time interaction vs. test-time compute}
\small
Under a fixed compute budget (measured by total tokens), gaining information through longer interaction with the environment can be more effective than solely deepening per-step reasoning. While longer chain-of-thought can improve local decision quality, interaction scaling offers a complementary and often more compute-efficient way to enable agents to adapt and explore over longer horizons.
\end{AIbox}

\vspace{-0.2cm}
\section{\methodname: Curriculum-Based Online RL for Scaling Interaction}
\vspace{-0.2cm}
How can we extend beyond prompting-based scaling to training agents that can effectively utilize interaction scaling? A natural starting point is to draw inspiration from current approaches for optimizing test-time compute~\citep{snell2025scaling,Qu2025OptimizingTC,DeepSeekAI2025DeepSeekR1IR} and extend these ideas to interactive settings. Specifically, we can run reinforcement learning (RL) with binary task rewards and longer task horizons. However, is this approach sufficient? We first describe the key challenges in learning to scale test-time interaction, and then develop our approach to address them via  curriculum learning.

\vspace{-0.2cm}
\subsection{Challenges of Training Agents with Fixed, Long Horizons} 
\label{sec:tti:challenges}
\vspace{-0.2cm}

\looseness=-1
A natural way to encourage the agent to learn to take more steps is to train at long horizons. To study this, we can run the simplest form of REINFORCE \citep{sutton1999policy} with binary rewards $R(\cdot)$, also known as online filtered behavior cloning (BC) or online STaR~\citep{Bai2024DigiRLTI, zelikman2022star}. Only successful trajectories are retained, and the agent is updated by maximizing the log-likelihood of actions conditioned on those high-reward rollouts:
\begin{align}
    \argmax_{\theta}~~ \mathbb{E}_{\mathcal{T}\sim \text{tasks}} \left[\mathbb{E}_{o_{0: h}, a_{0: h-1}\sim \pi(\cdot \mid \mathcal{T})}\left[ \left(\underbracket{\sum_{t=0}^{h-1} \log \pi_{\theta}(a_{t} \mid o_{\leq t}, \mathcal{T})}_{\text{likelihood of trajectory}}\right) \cdot \mathbbm{1}\left[\underbracket{R(o_{0:h}, \mathcal{T})}_{\text{reward}} =1 \right] \right] \right]
\end{align}
We use filtered BC as it is stable throughout training (no negative gradient~\citep{tajwar2024preference}), has been utilized previously for training agents~\citep{Bai2024DigiRLTI}, and is a good ``first-choice'' algorithm for studying the role of interaction scaling. We scale the agent's horizon on WebArena, varying $h \in \{5, 10, 20\}$. 
Smaller $h$ exposes the agent only to exploitative rollouts that succeed within allowed time steps, while larger $h$ also includes more exploratory rollouts. 
We use the non-test tasks for rollout. Details are in Appendix~\ref{app:filteredbc}.

As shown in Figure~\ref{fig:traindynamics} (left), the agent trained with $\horizon=5$ learns quickly, likely because on-policy RL is more sample-efficient at smaller horizons, but it also quickly overfits, and performance decreases with more training (x-axis)\footnote{We use the same training hyperparameters (Appendix~\ref{app:filteredbc}) across all settings for fair comparison. While reducing the learning rate can help reduce performance drop, it does not address the core issue with small horizons: fewer successful rollouts and shorter trajectories lead to significantly less training data.}. This  agent often terminates prematurely during evaluation despite being allowed   to interact for much longer time. Conversely, agents trained at longer horizons generally learn policies that are quite stochastic and learn significantly more slowly due to higher variance of the loss and credit assignment challenges due to longer horizons~\citeg{Wu2018VarianceRF,Zhao2011AnalysisAI,Castro2012PolicyGW}.  We manually inspect the trajectories and find that the $h=20$ agent tends to associate exploratory actions such as ``going back'' or ``trying random links'' with high rewards initially. 
Noisy credit assignment with $h=20$ slows learning, and only after several iterations do the agents begin to recover and produce more robust policies.
The impact of horizon is domain-dependent: in complex domains requiring exploration (e.g., CMS), long-horizon agents outperform, while in simpler settings (e.g., Reddit), performance differences are minimal. We also note that the total number of tokens generated per \textit{action} slightly decreases throughout training, indicating that the training does not incentivize increasing the length of reasoning directly.

Importantly, although the interaction length increases as expected for $h=20$ (Figure~\ref{fig:traindynamics}, right), noisy credit assignment and slower learning suggests that simply setting $h$ to be large is insufficient to learn to scale test-time interaction reliably. These observations motivate our method's core idea: rather than fixing the horizon throughout training, our approach aims to scale their interaction length dynamically.

\begin{figure}[t]
    \centering    \includegraphics[width=0.98\linewidth]{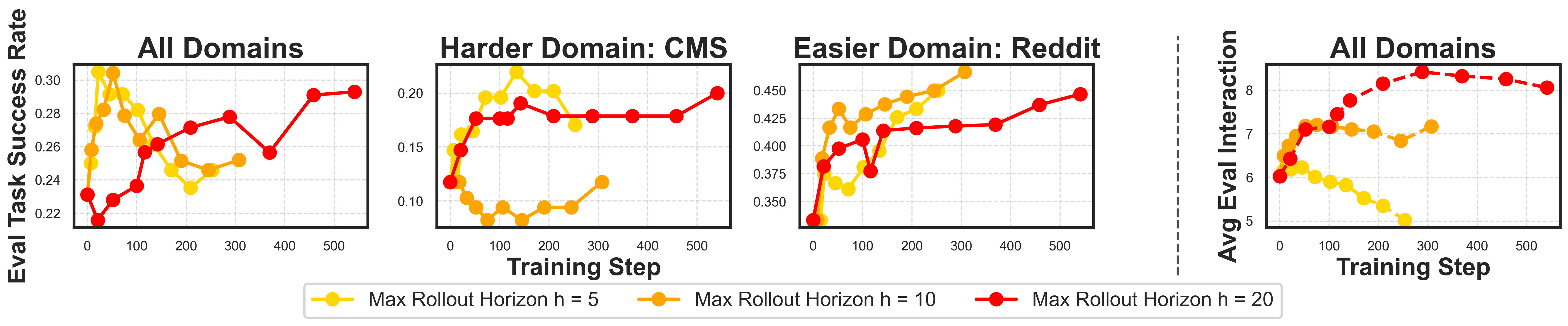}
    \vspace{-0.2cm}
    \caption{\footnotesize{\textbf{\textit{Online RL with different values of maximum interaction horizon.}} \textbf{Left:} success rates for different domains. ``Harder'' means generally lower success rate. \textbf{Right:} average rollout length ($\horizon_\mathrm{stop}$) on the evaluation set.}}
    \label{fig:traindynamics}
\end{figure}

\vspace{-0.2cm}
\subsection{Our Approach: Curriculum Over Interaction Horizon}
\label{sec:tti:schedule}
\vspace{-0.2cm}
To address these challenges, we propose \methodname{} (Test-Time Interaction), a curriculum-based online RL approach that trains the agent with short trajectories initially and gradually exposes it to longer ones.
Existing curriculum learning methods in RL~\citeg{Bengio2009CurriculumL, Narvekar2020CurriculumLF,Wang2021ASO, Matiisen2017TeacherStudentCL, Andrychowicz2017HindsightER} or web agents~\citep{qi2025webrltrainingllmweb,lai2024autowebglm} typically prioritize easier tasks before harder ones, often relying on predefined heuristics or external measures of task complexity. In contrast, we define curriculum progression in terms of the maximum number of steps an agent is allowed per trajectory. While interaction length can serve as a rough proxy for task difficulty, our approach does not require explicit labeling or estimation of task complexity.

\looseness=-1
\textbf{\emph{How do we design a curriculum over the interaction horizon $\horizon$?}} Ideally, the curriculum should allow the agent to first learn basic, ``atomic'' skills  to solve easier tasks,
then progressively tackle complex ones via skill chaining and exploration. To enable this kind of behavior, we explored two strategies: \textbf{(1)} a conservative, additive increase in $\horizon$ per iteration, giving the agent  sufficient time to solidify core task-solving skills; and \textbf{(2)} 
 a more aggressive, multiplicative increase, which assumes the  agent can quickly acquire the basic skills and benefit from earlier exposure to exploratory behavior. 
Formally, for iteration $i$:
{
\setlength{\abovedisplayskip}{6pt}
\setlength{\belowdisplayskip}{6pt}
\begin{align}
\footnotesize
h_i &:= \text{clip}(h_{\min} + i,\; h_{\max}) \quad \text{(Additive curriculum)} \\
h_i &:= \text{clip}(h_{\min} \cdot i,\; h_{\max}) \quad \text{(Multiplicative curriculum)}
\end{align}
}We store the rollouts in a replay buffer and assign higher weights to more recent trajectories. The full pseudocode for \methodname{} and  implementation details are provided in Appendix~\ref{app:tti}.

\begin{wraptable}{r}{0.22\textwidth}
\vspace{-0.5cm}\caption{\footnotesize{\emph{\textbf{Comparing various curricula.}} A multiplicative curriculum produces the best success rate.}}
\vspace{-0.2cm}
\label{table:schedule}
\Huge
  \centering
\resizebox{0.22\textwidth}{!}{\begin{tabular}{lcc}
\toprule
{\bf Schedule}   &{\bf Task SR (\%)}
\\ \toprule
Additive    & 29.50\\
Multiplicative   & 32.25 \\
\bottomrule
\end{tabular}}
\vspace{-0.5cm}
\end{wraptable}
\textbf{Empirical insights.} We instantiate these two strategies in   WebArena, using the non-test tasks for online training. We set $\horizon_{\min}$ to 10 and $\horizon_{\max}$ to 30,  and apply the schedules on top of   filtered BC. Evaluation results after 10   iterations are shown in Table~\ref{table:schedule}. Multiplicative curriculum outperforms the additive one, possibly because it exposes the agent to longer horizons early on and helps prevent it from overfitting prematurely to shortcut behaviors like always taking the shortest path. Based on these findings, we adopt the multiplicative curriculum as the default for \methodname{}. Table~\ref{table:schedule} further shows that even with limited data ($\sim$700 training tasks), \methodname{} outperforms fixed $h=20$ in Figure~\ref{fig:traindynamics} by nearly 3\%,  using 40\% fewer training steps over 10 iterations. 
 Next, we demonstrate that this advantage carries over to large-scale online RL training.

\begin{AIbox}{Takeaways: Curriculum training enables effective interaction scaling}
\small
Our approach, \methodname{}, gradually scales the horizon using a curriculum, leading to better performance than fixed-horizon baselines. The multiplicative schedule improves both learning efficiency and task success rate.
\end{AIbox}
\vspace{-0.2cm}
\section{Experiments: Scaling Up to Realistic Benchmarks}
\label{sec:exp}
\vspace{-0.2cm}
We now provide a comprehensive evaluation of \methodname{} in large-scale, realistic environments. More generally, we demonstrate the effectiveness of training models to make good use of test-time interaction.

\begin{wrapfigure}{r}{0.7\linewidth}
    \centering
    \vspace{-0.5cm}
    \includegraphics[width=0.99\linewidth]{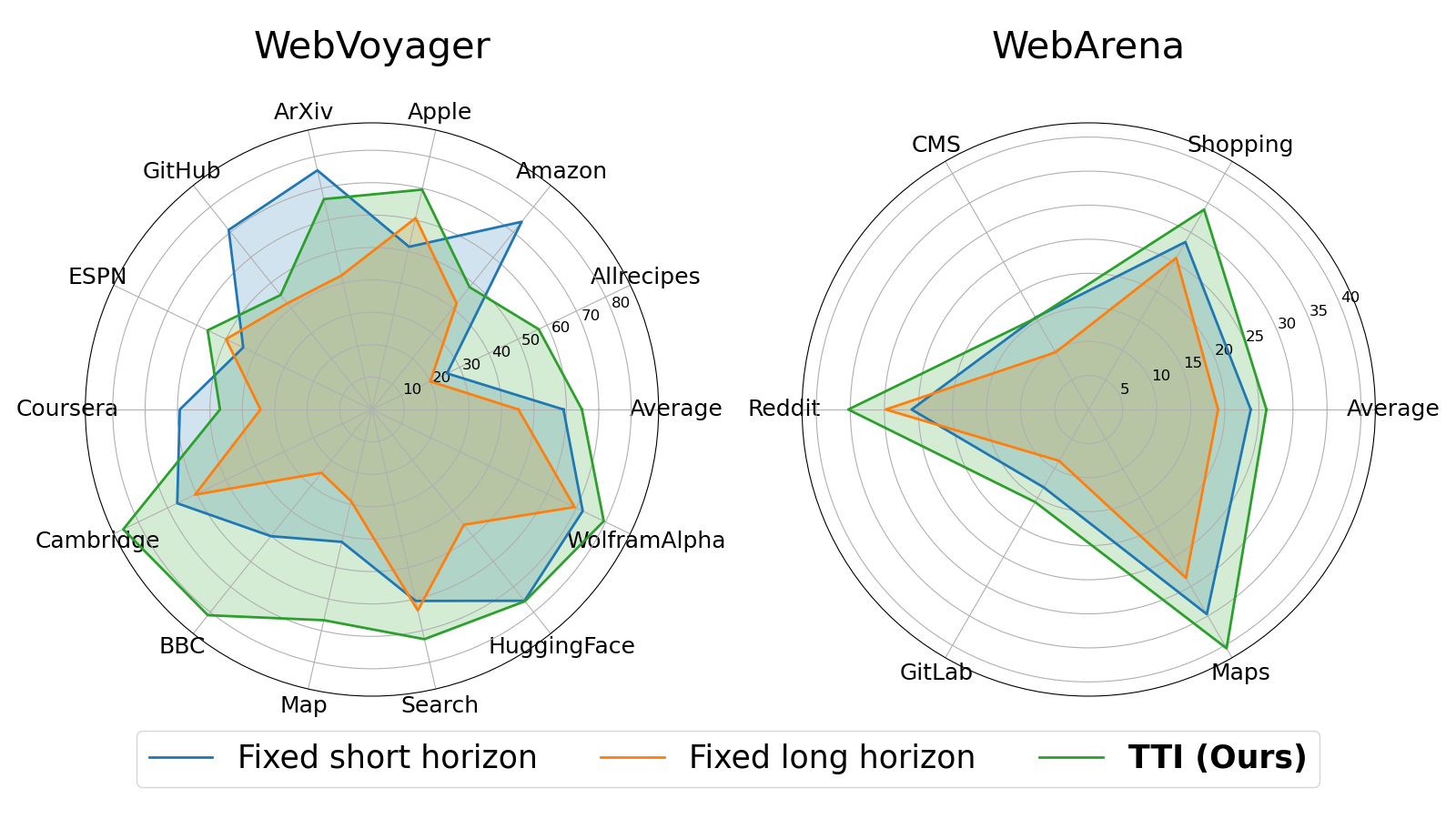}
    \vspace{-0.4cm}
    \caption{\footnotesize{\textbf{\emph{Summary of our main results on WebVoyager and WebArena benchmarks}.} \methodname{} consistently outperforms fixed-horizon training with both short and long horizons.}}
    \label{fig:result_summary}
    \vspace{-0.2cm}
\end{wrapfigure}
\textbf{Synthetic task generation and evaluation.} To enable large-scale training without training on the benchmark itself, we adopt synthetic task generation inspired by PAE~\citep{Zhou2024PAE}. Apart from prompting LLMs with seed examples or demos, we also deploy an exploratory agent to freely interact with websites and   propose more grounded, diverse tasks.  For evaluation, we leverage a prompting-based verifier  that uses action histories and screenshots to label rollouts, achieving 88.9\% accuracy compared to WebArena's ground-truth evaluator. Details and prompt templates are in Appendix~\ref{app:webvoyager:evalprompt}. We evaluate agents on (1) WebVoyager~\citep{WebVoyager} with 427 tasks across 13 domains (we replace Google search with Bing due to ReCaptcha issues); and (2)   full WebArena~\citep{zhou2024webarena} with 812 tasks. We choose these benchmarks as they are widely used~\citeg{openaioperator}.

\looseness=-1
\textbf{Training.} We obtain 128K synthetic tasks across diverse real-world domains  for WebVoyager and 11K tasks for WebArena's self-hosted domains (data released with the code). We train separate agents for each benchmark to avoid contaminating real-domain agents with synthetic environment data, as WebArena domains still have many differences from their real-world counterparts.
We use Gemma 3 12B~\citep{Kamath2025Gemma3T} as the base model, sampling 512 tasks per iteration for rollouts and updating with 512 successful on-policy trajectories. We apply a multiplicative curriculum with $h_{\min}=10$ and $h_{\max}=30$. We use vLLM~\citep{kwon2023efficient} to sample rollouts and use DeepSpeed Zero 3~\citep{Rasley2020DeepSpeedSO} with NVIDIA H100 GPUs for training. The evaluator is a Gemma 3 27B model, prompted to detect successful trajectories, which can then be used for the online filtered BC procedure. Other hyperparameters such as the number of iterations, learning rate, and the exact schedule of \methodname{} can be found in Appendix~\ref{app:webvoyager:hyperparam}.

\textbf{Comparisons.} We evaluate zero-shot Gemma 3 12B and approaches that utilize a fixed horizon with $h\in\{10, 30\}$. We also compare to closed-source agents (e.g., those based on GPT-4~\citep{gpt4} and Claude~\citep{claude3}), open-weight models trained on proprietary data (e.g., UI-TARS~\citep{qin2025ui}), and fully open-weight, open-data models (e.g., PAE~\citep{Zhou2024PAE}). A detailed list of the prior approaches can be found in the result tables.

\vspace{-0.2cm}
\subsection{WebVoyager Results and Analysis}
\vspace{-0.1cm}

\emph{\textbf{State-of-the-art open-weight, open-data performance.}} We report the overall task success rates (SR) on WebVoyager in Table~\ref{tab:webvoyager} and Figure~\ref{fig:result_summary} (left). The \methodname-trained Gemma 3 12B achieves an average SR of 64.8\%, setting a new state-of-the-art among open agents trained purely on public data. While previous methods such as UI-TARS achieves a strong SR of 84.8\%, they rely on private human-annotated data that remains inaccessible to the open-source community. In contrast, \methodname{} is trained entirely on   synthetic data and interestingly, this data is generated by the base model (Gemma 3 12B) itself, meaning that our training protocol implements a form of \emph{self-improvement}. This shows the practicality of our approach, but also highlights the need for future work on developing high-quality open data comparable to human-generated ones. \methodname{} also obtains the highest SR in 8 out of 13 domains, illustrating its efficacy.

\begin{table*}[t]
\centering
\caption{\footnotesize{\emph{\textbf{WebVoyager results.}} Training a Gemma3 12B model using \methodname{} attains the best performance among open-weight agents trained on open-source synthetic data in aggregate on the WebVoyager benchmark.} Baseline results are taken from \citet{Zhou2024PAE,qin2025ui}, and illustrate that \methodname{} improves over the best prior open-model, open-source data approach by 30\%.}
\Huge
\label{tab:webvoyager}
\resizebox{\textwidth}{!}{
\begin{tabular}{lcccccccccccccc}
\toprule
\textbf{Model} & \textbf{Average} & Allrecipes & Amazon & Apple & ArXiv & GitHub & ESPN & Coursera & Cambridge & BBC   &   Map & Search &  HuggingFace & WolframAlpha \\
\midrule
\multicolumn{15}{l}{\textbf{Proprietary Model}} \\

Claude 3.7   & 84.1  & - & - & - & - & - & - & - & - & - & - & - & - & -\\

Claude 3.5   & 50.5 & 50.0 & 68.3 & 60.4 & 46.5 & 58.5 & 27.3 & 78.6 & 86.0 & 36.6 & 58.5 & 30.2 & 44.2 & 66.7 \\
OpenAI CUA &87.0 & - & - & - & - & - & - & - & - & - & - & - & - & - \\

Agent E  & 73.1 & 71.1 & 70.7 & 74.4 & 62.8 & 82.9 & 77.3 & 85.7 & 81.4 & 73.8 & 87.8 & 90.7 & 81.0 & 95.7 \\
\midrule
\multicolumn{15}{l}{\textbf{Open Model, Proprietary Human-Annotated Data}} \\
UI-TARS-1.5  & 84.8 & - & - & - & - & - & - & - & - & - & - & - & - & - \\
\midrule
\multicolumn{15}{l}{\textbf{Open Model, Open Synthetic Data}} \\
LLaVa-34B SFT &22.2 & 6.8 &26.8 &23.3& 16.3& 4.9& 8.6& 26.8& 67.4 &16.7& 12.2& 23.3 &20.9 &38.1 \\
PAE-LLaVa-7B  & 22.3 & 14.3 & 37.5 & 17.5 & 19.0 & 14.6 & 0.0 & 33.3 & 52.4 & 18.6 & 22.5 & 23.3 & 19.0 & 24.4 \\
PAE-LLaVa-34B  & 33.0 & 22.7 & 53.7 & 38.5 & 25.6 & 14.6 & 13.6 & 42.9 & 74.4 & 39.0 & 22.0 & 18.6 & 25.6 & 42.9 \\
\cdashline{1-15}
Gemma 3 12B & 55.8 & 25.7 & 32.3 & 45.5 & {60.6} & 54.8 & \textbf{60.6} & 56.3 & 69.6 & 65.6 & 54.8 & \textbf{}\textbf{72.7 }& 66.7 & 61.1 \\
Fixed $h=10$   & 59.1 & 25.7 & \textbf{74.1} & 51.5 & \textbf{75.7} & \textbf{70.9} & 44.1 & \textbf{59.3} & 66.7 & 50.0 & 41.9 & 60.6 & 75.5 & 72.2 \\
Fixed $h=30$ &45.2&20.0&41.9&60.6&42.4&41.9&50.0&34.4&60.6&25.0&29.0&63.6&45.5&69.4\\
\textbf{\methodname{} (Ours)} & \textbf{64.8} & \textbf{57.1} & {48.3} & \textbf{69.6} & {66.6} & 45.2 & 56.3 & 46.9 & \textbf{85.2} & \textbf{81.2} & \textbf{66.7} & \textbf{72.7} & \textbf{75.7} & \textbf{79.4} \\
\bottomrule
\end{tabular}
}
\end{table*}

\begin{figure}[t]
    \centering
    \vspace{-0.3cm}
    \includegraphics[width=0.92\linewidth]{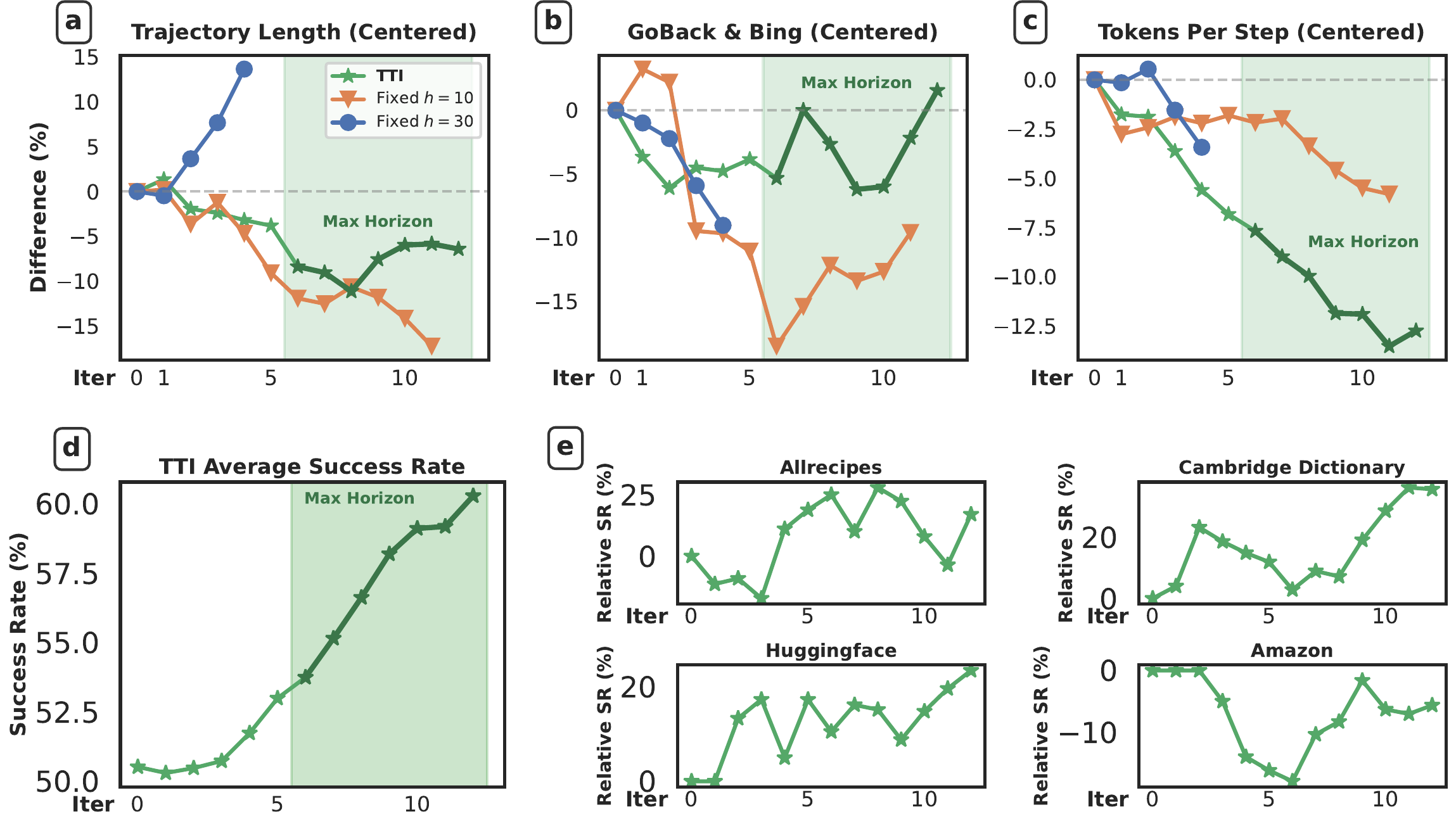}
   \caption{\footnotesize{\textbf{\emph{Dynamics of \methodname{}  during training.}} For \methodname{}, the green area represents the phase where the maximum allowed interaction horizon is the largest ($h=30$), per our multiplicative schedule. All results are evaluated on a held-out subset of WebVoyager, not on the training tasks.
   \textbf{a:} Average trajectory length, i.e. the average number of steps taken in a trajectory normalized by the average length at the first iteration (iteration 0).  \textbf{b:} Ratio of the sum of GoBack and Bing actions out of all actions normalized by the first iteration. 
   \textbf{c:} The average number of tokens each action uses, i.e., the average CoT length during agent reasoning. \textbf{d:} Average task success rates  for \methodname{} on the held-out validation set of tasks. \textbf{e:} Per-domain success rates  for \methodname{}. All results are evaluated on a held-out subset of WebVoyager tasks. Observe that \methodname{} learns to utilize more interaction steps and explores by calling \textit{GoBack} and \textit{Bing} actions once the maximum allowed horizon increases to peak value (i.e., in the green shaded area). The number of tokens appearing every step reduces for \methodname{} compared to the initialization, resulting in significantly shorter CoT compared to the run with $h=10$.}}
    \label{fig:webvoyager:train}
    \vspace{-0.3cm}
\end{figure}

\looseness=-1
\emph{\textbf{\methodname{} outperforms fixed-horizon via adaptive exploration.}} Table~\ref{tab:webvoyager} also shows that our curriculum approach outperforms fixed $h=10$ baseline by 5.7\% and fixed $h=30$ baseline by 19.6\% in average accuracy. 
To better understand the use of interaction within a training rollout, we plot the average number of interaction steps  
on a held-out validation set with 78 tasks in Figure~\ref{fig:webvoyager:train} (a). Note that the agent trained with $h = 10$ learns to continuously reduce the maximum number of steps it spends in a rollout, while $h = 30$ quickly drifts into aimless exploration and executes a larger number of steps pre-maturely in training, hindering performance. This aligns with our findings in Section~\ref{sec:tti:challenges} that simply running training at a longer horizon may not be sufficient for obtaining effective interaction scaling. In fact, we find that when training with \methodname{}, the interaction length of the agent's rollouts first decreases but then starts to increase as the maximum allowed horizon increases, indicating that an adaptive curriculum enables effective interaction scaling.

Figure~\ref{fig:webvoyager:train} (d) shows that the task success rate also grows over time and correlates with the expanding horizon.
While the average task success rates for \methodname{} are better, we observe notable per-domain differences. Figure~\ref{fig:webvoyager:train} (e) shows representative per-domain success rates. On domains like Allrecipes and Cambridge, \methodname{} significantly outperforms fixed-horizon and zero-shot approaches, improving success rates by 31.4\% and 15.6\%, respectively,  likely because these domains are highly information-dense and benefit from extended exploration enabled by adaptive interaction scaling. 
However, in domains like Amazon and GitHub, \methodname{} underperforms the baselines. We notice that the base model already has strong  knowledge about domain-specific terminologies (e.g., commit history, forks, stars) in these domains, likely because they are more prevalent than the others, resulting in high base performance. Inspecting the rollouts, we find that instead of using built-in filters and sorting, \methodname{} can engage in exploration behaviors such as initiating Bing searches or consulting external sites. This exposes the agent to noisy or distracting information, reducing task success. We discuss these cases in Section~\ref{sec:webvoyager:cases}.

\emph{\textbf{Learning dynamics of \methodname{}.}} To study how \methodname{} enhances the ``within-rollout'' exploration capabilities of the agent, we measure the number of \textit{GoBack} and \textit{Bing} actions over the course of training. \textit{GoBack} actions measure the number of retries the agent makes within an episode to get unstuck during exploration. \textit{Bing} actions correspond to the number of times the agent attempts to seek information by moving to \texttt{bing.com}. As shown in Figure~\ref{fig:webvoyager:train} (a, b, and d), the performance of \methodname{} improves substantially as the number of GoBack and Bing actions and the trajectory length grow.

Also note that the trajectory length and the numbers of \textit{GoBack} and \textit{Bing} actions begin to increase with \methodname{}, once the maximum allowed horizon length is increased as a part of the curriculum schedule (this regime is shown by the green shaded area in Figure~\ref{fig:webvoyager:train}). In contrast, these quantities continuously decrease over the course of training for the run with a lower number of maximum interaction steps ($h=10$). We also find that the trajectory length shoots up substantially for the run with $h=30$ and this correlates with worse performance. Finally, as shown in Figure~\ref{fig:webvoyager:train} (c) we also note that as the agent's trajectory grows longer with \methodname{}, the number of tokens appearing in per-step reasoning actually becomes smaller. This implies that our agent is automatically learning to tradeoff interaction for per-step compute in order to attain higher performance, and perhaps prevents any issues with overthinking.

\vspace{-0.35cm}
\subsection{Case Studies: Strengths and Failure Modes}
\label{sec:webvoyager:cases}
\vspace{-0.2cm}

We conduct detailed case studies to analyze how \methodname{} behaves across tasks and domains. These cases highlight both the strengths and remaining limitations of our approach.

\underline{\textbf{Strength: Effective exploration in complex tasks (example visualized in Appendix~\ref{app:case:strength1}).}}
For complex, exploratory tasks that require information retrieval, \methodname{} trains the agent to  extend its interaction horizon through search and backtracking, thus gathering and comparing   information before making decisions. For instance, when tasked to find the baking temperature of an apple pie recipe with 4+ stars and 50+ reviews, our agent first selects a recipe but encounters a pop-up it cannot dismiss due to backend issues. It then tries another recipe but finds no baking temperature. Returning to the listing again, it correctly identifies  one that meets all criteria. We also observe that such behaviors emerge progressively. In early training with shorter horizon,  \methodname-agent  tends to stick to the first recipe it finds, keeps retrying it and saying \textcolor{ForestGreen}{\textit{``I remember seeing one with 613 ratings earlier''}} instead of seeking alternatives. Only after the training horizon becomes longer does it learn to explore and backtrack. 
This illustrates that when \methodname{} runs a curriculum over interaction length, it teaches agents to adjust their horizon \textit{within} a task and shift from exploitation to exploration. In contrast, training with a fixed short horizon can make it difficult   to develop such exploratory behaviors. 

\looseness=-1
\underline{\textbf{Strength: Strategic exploitation in simple tasks (Appendix~\ref{app:case:strength2}).}}
For simpler tasks with clear, deterministic paths (e.g., form filling or direct lookups), \methodname-agent completes tasks efficiently without over-exploration. For example, when instructed to find the ``top trending open-source project on machine learning'' in GitHub, the agent goes directly to the Open-Source menu, selects the Trending tab, and performs search. This shows that \methodname{} balances exploration and exploitation based on task context. 

Despite these strengths, we also observe characteristic failure modes that point to areas for improvement and may partly explain the agent's lower performance on domains like GitHub.

\underline{\textbf{Failure mode: over-reliance on resets (Appendix~\ref{app:case:fail1}).}} When an action fails, our agent can reset the task by returning to the Bing search page rather than attempting recovery within the target domain. This suggests the agent treats search as a universal fallback, even when  more domain-specific actions (e.g., revisiting menus, refining filters) would be more effective. We also observe repeated resets within the same trajectory, indicating a lack of adaptive error recovery. While agents can extend horizons through both resetting and backtracking, the latter is often more natural. This highlights an area where \methodname{} could improve by guiding exploration more systematically and enforcing structure. We believe that using dense rewards~\citep{Qu2025OptimizingTC,Setlur2024RewardingPS} or running full multi-turn RL with value functions~\citep{zhou2024archer} may address this issue.

\underline{\textbf{Failure mode: limited self-verification (Appendix~\ref{app:case:fail2}).}} We also observe that the agent can fail to verify its actions against the task goal, especially in the last step. In one case, the agent identifies a 2021 GitHub repository for a task requiring one from 2022.  While it explicitly acknowledges the mismatch, \textcolor{ForestGreen}{\textit{``It was created in 2021, not 2022, so it doesn't meet the criteria''}}, it still submits it as the answer. This implies limited self-verification ability and could be mitigated by longer, more deliberate per-step reasoning. An important next step is to enrich  \methodname{} with the cognitive behaviors that enable per-step reasoning~\citep{Gandhi2025CognitiveBT}.

\begin{table}[t!]
\centering
\caption{\footnotesize{\textbf{\textit{Full WebArena results.}} For proprietary agents, we include the top 8 from the official leaderboard. Observe that TTI attains the best performance on an average, with especially better performance on the CMS domain.}}
\label{table:webarenafull}
\resizebox{\textwidth}{!}{
\begin{tabular}{lllcccccc}
\toprule
&\textbf{Method} &\textbf{Backbone}&\textbf{Average} & \textbf{Shopping} & \textbf{CMS} & \textbf{Reddit} & \textbf{GitLab} & \textbf{Maps} \\
\toprule
\textbf{Proprietary-Based}&IBM CUGA~\citep{Marreed2025TowardsEC}&-&61.7&-&-&-&-&-\\
&OpenAI CUA~\citep{openaioperator}&-&58.1&-&-&-&-&-\\
&Jace AI~\citep{jace}&-&57.1&-&-&-&-&-\\
&ScribeAgent~\citep{Shen2024ScribeAgentTS}& GPT-4o + Qwen2.5 32B&{53.0}&45.8&{37.9}&{73.7}&{59.7}&{56.3}\\
&AgentSymbiotic~\citep{Zhang2025SymbioticCF}& Claude 3.5 + Llama 3.1 8B  &48.5& 48.7 &41.2&63.2& 47.2 &57.8 \\
&Learn-by-Interact~\citep{Su2025LearnbyinteractAD}& Claude 3.5 Sonnet&48&-&-&-&-&-\\

&AgentOccam-Judge~\citep{AgentOccam}&GPT-4&45.7 &43.3& {46.2}&67.0& 38.9 &52.3\\
&WebPilot~\citep{zhang2024webpilot}&GPT-4o &37.2& 36.9 &24.7	&65.1	&39.4	&	33.9	\\
\midrule
\textbf{Fully Open-Source}
&Learn-by-Interact~\citep{Su2025LearnbyinteractAD}& Codestral 22B&24.2&-&-&-&-&-\\
\textbf{(Self-Improvement)}&AgentTrek~\citep{Xu2024AgentTrekAT}&Qwen2.5 32B & 22.4&-&-&-&-&-\\
&AutoWebGLM~\citep{lai2024autowebglm}& ChatGLM3 6B&18.2&-&-&-&-&-\\

& NNetnav~\citep{Murty2024NNetNavUL} & Llama 3.1 8B&7.2&7.4&4.2&0&0&28.5\\

\cdashline{1-9}
 &Zero-Shot Baseline&Gemma 3 12B&18.3&26.7&8.7&30.9&5.5&27.7\\
&Fixed $h=10$ &Gemma 3 12B&23.8&28.4&\textbf{15.6}&26.0&13.2&34.7\\
&Fixed $h=30$ &Gemma 3 12B&19.0&25.7&{9.7}&29.8&8.7&28.57\\

&\textbf{\methodname{} (Ours)} &Gemma 3 12B&\textbf{26.1}&\textbf{33.9}&15.5&\textbf{35.3}&\textbf{15.7}&\textbf{40.5}\\

\bottomrule
\end{tabular}}
\end{table}
\vspace{-0.2cm}
\subsection{WebArena Results and Analysis}

We further assess \methodname{}  on the full WebArena~\citep{zhou2024webarena} (we only use the prompting-based verifier for training, but use the original benchmark evaluators for evaluation).  As shown in Table~\ref{table:webarenafull}  and Figure~\ref{fig:result_summary} (right), \methodname{} obtains   the highest performance among open-source agents trained entirely using a self-improvement procedure on self-generated data, without relying on proprietary models for task completion or distillation. While \methodname{} improves over the zero-shot baseline by 7.8\%, the gains are smaller than on WebVoyager, possibly because: (1) WebArena tasks are more complex, as reflected in lower accuracies even for proprietary models, leading to fewer successful rollouts per iteration and slower learning; (2) Agents occasionally attempt actions that are valid in real-world domains but invalid in WebArena counterparts (for example, search works reliably on Reddit but fails in WebArena's Postmill due to environment bugs).
More experiment details are in Appendix~\ref{app:fullwebarena}.

\begin{wrapfigure}{r}{0.35\textwidth}
\centering
        \vspace{-0.5cm}
\includegraphics[width=0.99\linewidth]{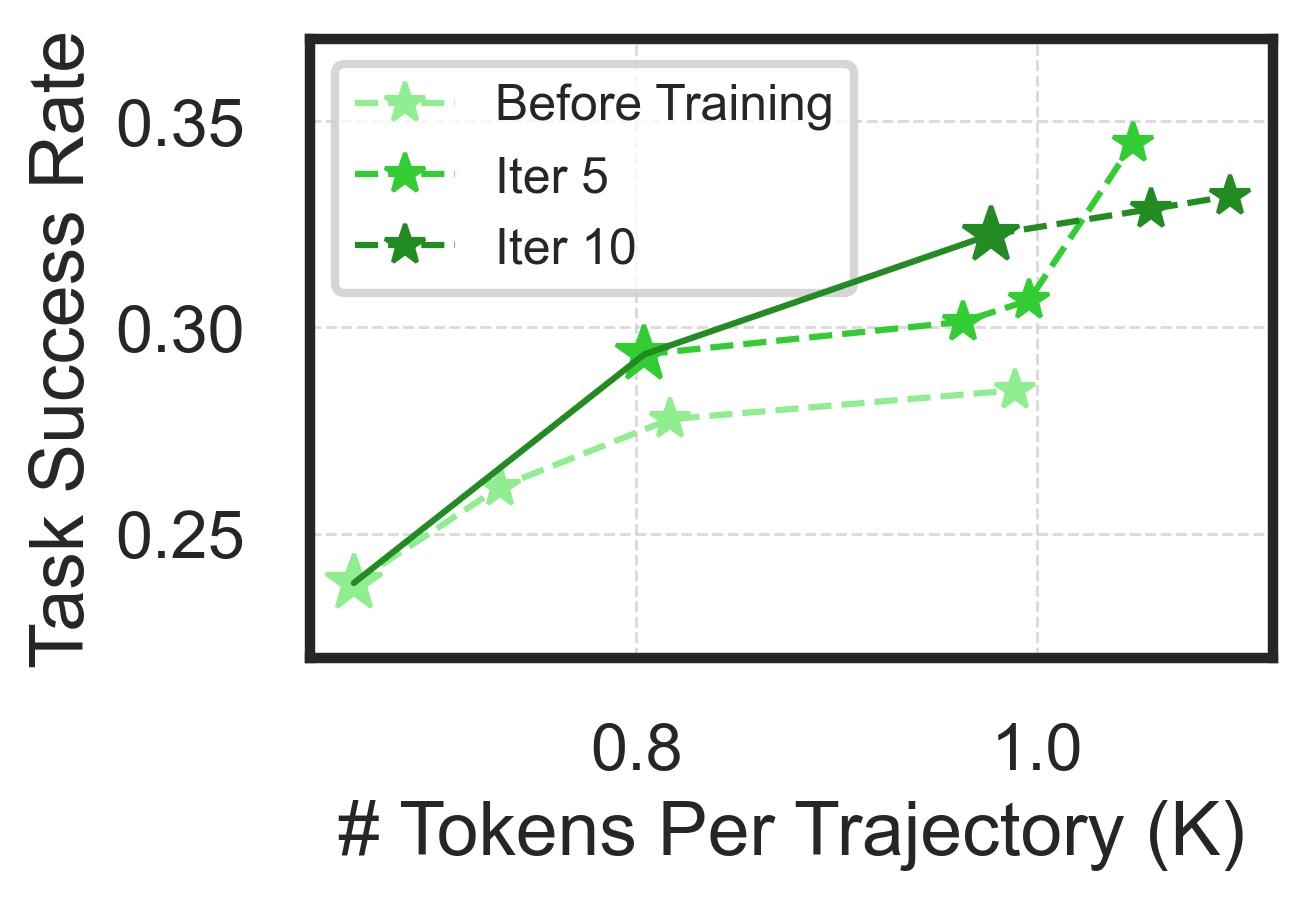}
\vspace{-0.8cm}
    \caption{\footnotesize{\textit{\textbf{We apply test-time re-checks to  \methodname{} checkpoints.}} Note that inference-time re-checking improves performance on all \methodname{} checkpoints, and more importantly, training further with TTI improves performance more than inference-time re-checks as expected.}}
    \vspace{-0.3cm}
    \label{fig:ckpt}
\end{wrapfigure}
\textbf{\underline{Further scaling.}} While \methodname{} equips agents with the ability to adjust their interaction horizon during deployment, an open question remains: Can we further amplify performance by combining TTI with inference-time interaction scaling techniques such as re-checking as discussed in Section~\ref{sec:paradigm}? To explore this, we apply the ``check-again'' strategy (Section~\ref{sec:paradigm}) to intermediate \methodname{} checkpoints. Due to the high evaluation cost associated with evaluating on full WebVoyager or WebArena, we leverage the WebArena subset  checkpoints obtained   in Section~\ref{sec:tti:schedule}. 

As shown in Figure~\ref{fig:ckpt}, applying re-checking on top of \methodname{}  improves task success   across various training stages. The benefits are more obvious in the early stages of training, when the agent has a stronger bias to terminate prematurely.
As training progresses, \methodname{} encourages longer interaction traces that naturally incorporate behaviors like re-checking, reducing the added benefit of explicit re-checks. Nonetheless, even in later stages, re-checking does continue to provide modest gains in performance, serving as a  safety-check for well-trained agents.
\vspace{-0.2cm}
\section{Conclusion and Future Work}
\vspace{-0.2cm}
In this work, we introduced interaction scaling as a new dimension of test-time scaling for interactive agents. 
Through empirical studies on web agents, we validate that interaction scaling enables agents to explore and adapt dynamically, significantly improving task performance. Despite the promising results, there are a number of avenues for future work which we list below. 

\vspace{-0.4cm}
\begin{itemize}[itemsep=4pt]
    \item \textbf{Extension to other agentic problem domains.} While we only study the effects of interaction scaling on web environments, we believe that this procedure is likely going to be even more effective in domains that admit more stochasticity and uncertainty, such as robotic control or open-world compute use problems. In these settings, solving tasks would require gather information first before attempting to solve the task. By utilizing interaction as an effective tool for information gathering, we can attain much better performance.
    \item \textbf{Balancing the tradeoff between thinking more and acting more.} While we illustrate that spending a given amount of total token budget on acting for longer can be more effective than reasoning for longer, an interesting open question is how we should incentivize RL training to best balance thinking and acting during the process of learning. In our experiments, we observed an increased preference towards acting longer, with the number of tokens in per-step reasoning decreasing  compared to the initialization. While this results in better use of total overall test-time compute given our results in Section~\ref{sec:paradigm}, it is unclear as to what a general thumb rule and training procedure should be to attain the best tradeoff between acting longer and thinking more.
    \item \textbf{Improved RL algorithms for powering interaction scaling.} The RL approach we utilize in this work is extremely simple as it corresponds to online filtered behavior cloning (BC). An immediate next promising direction is to extend \methodname{} to utilize negative gradients~\citep{tajwar2024preference} via GRPO~\citep{Shao2024DeepSeekMathPT} or PPO~\citep{schulman2017ppo}. However, stochasticity and, even non-stationarity, in interactive agent environments suggests that RL algorithms that train value functions~\citep{Bai2025DigiQLQ,zhou2024archer} are likely to be more successful at effective credit assignment as the task horizon is scaled further. In addition, we will need to address challenges pertaining to memory and context length, that is very likely to overflow as we scale horizon even further. Tackling any of these challenges would be exciting for future work.
\end{itemize}

% \textbf{Limitations and future work. }
% Our experiments are limited to  web environments; extending this method to other domains like robotics or open-world games requires further exploration. Besides, scaling interaction steps increases computational costs during both inference and training. Although our adaptive scheduling helps, more efficient handling of long interactions is needed. In addition, our training relies on simple behavior cloning; future work could incorporate more advanced RL methods like PPO~\citep{Schulman2017ProximalPO}, GRPO~\citep{DeepSeekAI2025DeepSeekR1IR} to improve  performance. Lastly, due to high compute cost, we only ran the full benchmark once per setting, limiting the ability to quantify variance from policy, environment, and evaluator stochasticity. Future work should explore multiple runs or  more robust evaluation.

\vspace{-0.2cm}
\section*{Acknowledgments}
\vspace{-0.2cm}
We thank {Jiayi Pan, Shanda Li, and Yuxiao Qu} for feedback and informative discussions on an earlier version of this paper. This work was supported in part by the National Science Foundation grants IIS1705121, IIS1838017, IIS2046613, IIS2112471, the Office of Naval Research under N00014-24-1-2206, and funding from Meta, Morgan Stanley, Amazon, Google, and Scribe. We specially thank Scribe, The AGI company, and CMU FLAME Center (Orchard cluster) for providing a part of the GPU resources that supported this work. Any opinions, findings and recommendations expressed in this material are those of the author(s) and do not necessarily reflect the views of any of these funding agencies or employers.

\newpage
\bibliographystyle{unsrtnat}
\bibliography{reference}

\newpage

%%%%%%%%%%%%%%%%%%%%%%%%%%%%%%%%%%%%%%%%%%%%%%%%%%%%%%%%%%%%

\appendix
\onecolumn

\part*{Appendices}

% \vspace{-0.2cm}
\section{Broader Impact}
% \vspace{-0.2cm}
This work contributes to the development of more adaptive and capable AI agents by introducing a new test-time scaling dimension focused on interaction rather than per-step reasoning alone. While this approach improves robustness and generalization in open-ended environments, it also raises important considerations. Increased agent autonomy can amplify both the benefits and risks of deployment in real-world systems. Moreover, agents capable of richer behaviors could be applied to sensitive domains (e.g., customer service, education, or automation workflows) where unintended actions could have large impacts. We encourage future work to consider ethical safeguards, interpretability tools, and human-in-the-loop designs when deploying interaction-scaled agents. Our experiments are conducted entirely in simulated environments, and we hope this work inspires further research on controllable and trustworthy agent behavior under realistic constraints.

% \vspace{-0.2cm}
\section{Observation Space Design}
\label{app:sec:observation}
% \vspace{-0.2cm}
We use  the screenshot accompanied with the web page's accessibility tree as our main observation. We study two versions of accessibility tree. \textbf{Rich accessibility tree} is modified from the WebArena code and looks like:

[21]: RootWebArea 'Dashboard / Magento Admin' focused: True;	[0]: link 'Magento Admin Panel';	[1]: link 'DASHBOARD';	[2]: link 'SALES';	[3]: link 'CATALOG';	[4]: link 'CUSTOMERS';	[5]: link 'MARKETING';	[6]: link 'CONTENT';	[7]: link 'REPORTS';	[8]: link 'STORES';	[22]: link 'SYSTEM';	[23]: link 'FIND PARTNERS \& EXTENSIONS';	[24]: heading 'Dashboard';	[9]: link 'admin';	[10]: link '';	[25]: StaticText 'Scope:';	[12]: button 'All Store Views' hasPopup: menu;	[13]: link 'What is this?';	[14]: button 'Reload Data'...

\textbf{Simple accessibility tree} is modified from the PAE code and looks like:

[1]: "Dashboard";	[2]: "Sales";	[3]: "Catalog";	[4]: "Customers";	[5]: "Marketing";	[6]: "Content";	[7]: "Reports";	[8]: "Stores";	[9]: "admin";	[12]: <button> "All Store Views";	[13]: "What is this?";	[14]: <button> "Reload Data";	[15]: "Go to Advanced Reporting";	[16]: "here";...

Rich tree contains more details such as the HTML tag and attributes like \texttt{required}, \texttt{hasPopup} compared to simple tree. However,  it is much longer than simple tree and hence harder to optimize due to the increased context length.

\begin{algorithm}[h!]
    \caption{\methodname{}: Filtered Behavior Cloning with Interaction Scheduling}
    \label{alg:tti}
    \begin{algorithmic}[1]
    \State \textbf{Input:} Agent policy $\pi_{\theta}$, Evaluator $\mathcal{R}$, Environment $\mathcal{E}$, Learning rate $\alpha$, Replay buffer $\mathcal{D}$, Interaction scheduler hyperparameters $h_{\min}$, $h_{\max}$
    \State Initialize policy $\pi_{\theta}$ from pretrained model
    \State Initialize replay buffer $\mathcal{D} \leftarrow \{\}$
    \For{each episode $i$}
        \State Set interaction horizon $h_i \leftarrow \text{get\_schedule}(i, h_{\min}, h_{\max})$
        \For{each rollout to collect}
            \State Initialize environment: $s_0 \sim \mathcal{E}$
            \For{each $t$ in $[1, h_i]$}
                \State Observe current state $s_t$
                \State Predict action $\hat{a}_t \leftarrow \pi_{\theta}(s_t)$
                \State Execute action $\hat{a}_t$ in environment
                \State Observe next state $s_{t+1}$
                \If{episode done}
                    \State Compute reward $r_t \leftarrow \mathcal{R}(s_t, \hat{a}_t)$
                \Else
                    \State $r_t \leftarrow 0$
                \EndIf
                \State Store transition: $\mathcal{D} \leftarrow \mathcal{D} \cup \{(s_t, \hat{a}_t, r_t, s_{t+1})\}$
            \EndFor
        \EndFor
        \For{sample successful trajectory in $\mathcal{D}$}
            \For{$t = 1$ to $h_{\text{stop}}$}
                \State Accumulate loss: $L(\theta) \leftarrow L(\theta) + \text{CrossEntropy}(\pi_{\theta}(s_t), \hat{a}_t)$
            \EndFor
        \EndFor
        \State Update policy: $\theta \leftarrow \theta - \alpha \nabla_{\theta} L(\theta)$
    \EndFor
    \end{algorithmic}
\end{algorithm}

\section{\methodname{} Implementation}
\label{app:tti}

We provide the pseudocode in Algorithm~\ref{alg:tti}. For the replay buffer, to encourage the agent to learn from more recent examples, we assign weights based on recency when sampling rollouts to update the agent: for the $k$-th trajectory added to the buffer, its weight is $\frac{k}{|\mathcal{D}|}$.

\section{WebArena   Experiments}
\subsection{WebArena Agent Prompt}
\label{app:prelim:prompt}
\begin{tcolorbox}[
  colback=white,
  colframe=darkgray,
  title=CoT Prompt,
  fonttitle=\bfseries\centering,
  top=2mm,
  bottom=2mm,
  arc=3mm,
  breakable
]
Imagine you are an agent browsing the web, just like humans. Now you need to complete a task. In each iteration, you will receive an observation that includes the accessibility tree of the webpage and a screenshot of the current viewpoint. The accessbility tree contains information about the web elements and their properties. The screenshot will feature numerical labels placed in the TOP LEFT corner of web elements in th current viewpoint. Carefully analyze the webpage information to identify the numerical label corresponding to the web element that requires interaction, then follow the guidelines and choose one of the following actions:

1. Click a web element.

2. Delete existing content in a textbox and then type content.

3. Scroll up or down the whole window.

4. Go back, returning to the previous webpage.

5. Answer. This action should only be chosen when all questions in the task have been solved.

Correspondingly, action should STRICTLY follow the format specified by one of the following lines:

Click [numerical\_label]

Type [numerical\_label] [content]

Scroll [up/down]

GoBack

ANSWER [content]

Some examples are:

Click [8]

Type [22] [Boston]

Scroll [down]

ANSWER [06516]

Key guidelines you MUST follow:

* Action guidelines *

- Use either screenshot or accessibility tree to obtain the numerical\_label. Sometimes the accessibility tree captures more elements than the screenshot. It's safe to select these elements without scrolling

- For text input, use Type action directly (no need to click first). All existing texts in the textbox will be deleted automatically before typing

- Preserve text inside quotation marks exactly as provided by user

- You must not repeat the same actions over and over again. If the same action doesn't work, try alternative approaches

- Use ANSWER only after completing ALL task requirements

- Wrap content for Type and ANSWER with square brackets `[]`

- Do not add quotation marks for search queries

* Web navigation hints *

\{hint\}

Your reply should strictly follow the format:

Thought: Your reasoning trace. A good practice is to summarize information on the current web page that are relevant to the task goal, then generate a high-level plan that contains the sequence of actions you probably need to take

Action: Based on this reasoning, identify the single most optimal action. You should output it in the format specified above (under "STRICTLY follow the format")

After each action, you'll receive a new observation. Proceed until task completion. Now solve the following task.

Task: \{task\_goal\}

Current URL: \{url\}

Screenshot of current viewpoint: attached

Accessibility tree of current viewpoint:
\{accessibility\_tree\}
\end{tcolorbox}

Beyond the above CoT prompt, we also tried using a more complex prompt for the thought process. However, this does not lead to significant gain in downstream accuracy (see Table~\ref{app:table:cotchoice}), but it could increase training and inference cost, so we did not use it in the end.
\begin{tcolorbox}[
  colback=white,
  colframe=darkgray,
  title=Complex Prompt,
  fonttitle=\bfseries\centering,
  top=2mm,
  bottom=2mm,
  arc=3mm,
  breakable
]
Thought: You must analyze the current webpage thoroughly to guide your decision-making. Show your reasoning through these steps:

- Summarization: Begin by understanding the page context - identify what type of page you're on (search results, form, article, etc.) and how it relates to your objective. Summarize important information on the webpage that might be relevant to task completion. Especially when the task requires to return some answers to a specific question, you should note down intermediate information that helps generate the answer.

- Planning: Generate a checklist of subtasks required for completion and cross-out the subtasks you've completed. Identify the next logical subtask.

- Verification: Verify all information you've entered so far. Check that your inputs match requirements in terms of spelling and format (you should not change the user-specified information, even if there're grammar errors). Verify if any selections for dropdown items align with the task objective. Identify if there're necessary fields that have not been filled in. Note that if the last few steps are repeating the same action, there must be missing or incorrect information.

- Backtracking: If the task requires exploring multiple webpages (e.g., orders, posts, item pages, etc) to find out an answer, consider if you need to issue GoBack and return to the previous web page.

- Candidate Generation: After all the above reasoning, list the most relevant possible actions, evaluate pros and cons of each action, and finally select the most effective action to progress task.

Action: Choose ONE of the following action formats:

- Click [numerical\_label] - Click a specific element

- Type [numerical\_label] [content] - Input text into a field

- Scroll [up/down] - Navigate the page vertically

- GoBack - Return to previous webpage

- ANSWER [content] - Provide final answer when task is complete
\end{tcolorbox}

\subsection{WebArena Domain-Specific Prompts}

Below are the content replacing ``\{hint\}'' in the general prompt.

\begin{tcolorbox}[
  colback=white,
  colframe=darkgray,
  title=General Hint,
  fonttitle=\bfseries\centering,
  top=2mm,
  bottom=2mm,
  arc=3mm,
  breakable
]
- Always save progress through appropriate buttons (Save, Submit, Post, etc.)

- Always remember to interact with dropdown options after expanding

- Clear filters before setting new ones
\end{tcolorbox}

\begin{tcolorbox}[
  colback=white,
  colframe=darkgray,
  title=Reddit,
  fonttitle=\bfseries\centering,
  top=2mm,
  bottom=2mm,
  arc=3mm,
  breakable
]
- Always save progress through appropriate buttons (Save, Submit, Post, etc.)

- Always remember to interact with dropdown options after expanding

- Pay attention to words like "latest", "newest", "hottest" in the task objective, which require clicking the dropdown menu and select "New" or "Top" with the correct time range

- When selecting a subforum, you can either browse the dropdown menu in the "Submit" page or navigate to "Forums" and check all subforums by clicking on "Next" to go over all pages. You must try to find a subforum that exactly matches your query. If there's no exact match, pick the most relevant one, ideally the subforum is about objects or locations contained in the given objective

- "Trending" means "hot"

- To find out all posts or replies from a user, click the user name and then click "Submissions" or "Comments"
\end{tcolorbox}

\begin{tcolorbox}[
  colback=white,
  colframe=darkgray,
  title=CMS,
  fonttitle=\bfseries\centering,
  top=2mm,
  bottom=2mm,
  arc=3mm,
  breakable
]
- Always save progress through appropriate buttons (Save, Submit, Post, etc.)

- Always remember to interact with dropdown options after expanding

- Clear filters before setting new ones

- Use date format: month/day/year (e.g., 1/1/16, 12/31/24)

- When searching phone numbers, remove the country code

- When searching product name, use single but not plural form

- When the web page contains a table, aggregate the rows with the same item
\end{tcolorbox}

\begin{tcolorbox}[
  colback=white,
  colframe=darkgray,
  title=Shopping,
  fonttitle=\bfseries\centering,
  top=2mm,
  bottom=2mm,
  arc=3mm,
  breakable
]
- Always save progress through appropriate buttons (Save, Submit, Post, etc.)

- Always remember to interact with dropdown options after expanding

- Sort items by price by clicking the dropdown menu and set descending/ascending direction

- When searching product name, use single but not plural form

- If the objective requires only finding an item, stop at the item page without adding to cart

- To find out the quality of a product, search the item, click on review, and inspect its review

- Click "Page Next" to iterate over all orders

- Since there's no way to filter order history, click "View Order" for every order within a date range and inspect individually. If the condition is not met, go back
\end{tcolorbox}

\begin{tcolorbox}[
  colback=white,
  colframe=darkgray,
  title=GitLab,
  fonttitle=\bfseries\centering,
  top=2mm,
  bottom=2mm,
  arc=3mm,
  breakable
]
- Always save progress through appropriate buttons (Save, Submit, Post, etc.)

- Always remember to interact with dropdown options after expanding

- Clear filters before setting new ones

- When searching a repo in gitlab, type only the project name after "/" in the search box
\end{tcolorbox}

\begin{tcolorbox}[
  colback=white,
  colframe=darkgray,
  title=Map,
  fonttitle=\bfseries\centering,
  top=2mm,
  bottom=2mm,
  arc=3mm,
  breakable
]
- Always remember to interact with dropdown options after expanding

- When searching for a place, remove prepositions like in/on/by/at. For example, use "starbucks, craig street" instead of "starbucks on craig street". Put the city name at the end

- When there is no results shown up after search, rephrase the address and try again

- To find direction between two points, after entering the from and to addresses, select the correct transportation (foot/bicycle/car) before clicking "Go"

- When the given location is not a geological address, use your knowledge to infer the address
\end{tcolorbox}

\subsection{CoT Experiments for Base Agent}
\label{app:prelim:cot}
To enable efficient rollout collection, we spin up multiple Docker containers on a single GPU according to the official WebArena repository. We use the vLLM~\citep{kwon2023efficient} engine for inference and apply the following inference hyperparameters for most of our experiments.
\begin{itemize}
    \item max\_new\_tokens: 1024
    \item max\_attached\_imgs: 4
    \item temperature: 1
    \item top\_p: 0.95
\end{itemize}

We randomly subsample 62  test tasks for analysis purposes. Below are the results of zero-shot agent vs CoT prompting. ``CoT'' uses the   ``General Prompt'' in Section~\ref{app:prelim:prompt}. ``Complex CoT'' uses the   ``Complex Prompt'' in Section~\ref{app:prelim:prompt}.

\begin{table}[h!]
\centering
\caption{Base agent results averaged over 3 runs on WebArena subset.}
\label{app:table:cotchoice}
  \centering
\resizebox{0.25\textwidth}{!}{\begin{tabular}{lcc}
\toprule
{\bf Prompt}   &{\bf Task SR (\%)}
\\ \toprule
Action Only   & 14.76\\
 CoT   & 23.81 \\
 Complex CoT & 23.33\\
\bottomrule
\end{tabular}}
\end{table}

\subsection{Scaling Trade-off Experiments}
\label{app:prelim:scaling}
\paragraph{``Check-again'' for interaction scaling.}
After the agent outputs the task-stop signal, we append the following prompts to the observation to induce it to check again.
\begin{tcolorbox}[
  colback=white,
  colframe=darkgray,
  title=Check-Again Prompt,
  fonttitle=\bfseries\centering,
  top=2mm,
  bottom=2mm,
  arc=3mm,
  breakable
]
Important: You returned an answer in the last step. Let's pause, check the web page, and think again. If you still think the task is finished, double-check your answer, revise it if need, and return a final answer. If not, continue the task. Your output should still be in the same ``Thought:...Action:...'' format. 
\end{tcolorbox}

When applying multiple re-checks, we slightly vary the prompts such as `\textit{`Before you finalize the answer, re-evaluate it in terms of the current web page---what exactly supports or contradicts it?''} or \textit{``Why do I believe this answer is correct? What on the page justifies it? Could an alternative answer be better?''}
Please refer to the code base for the exact prompt used.

\paragraph{Per-step budget forcing.}
Following \citep{Qu2025OptimizingTC}, we use the phrases below to induce longer per-step thinking. The phrases are different to ensure that the model does not run into the scenario of endless
repeating a phrase.
\begin{itemize}
    \item First time: Wait, let me think deeper.
    \item Second time: But let me double-check.
    \item Third time: But hold on.
\end{itemize}

\paragraph{Per-step best-of-$n$.}
We tried both selecting by log likelihood and majority voting, with the latter showing slightly better results.

\paragraph{Additional results for combined scaling.} Beyond evaluating each scaling method separately, we also tried combining methods along different axes.

\begin{table}[h!] 
\caption{\small Comparing different inference-time prompting strategies. Results averaged over 3 runs on WebArena subset. All methods are   applied once.}
\label{app:table:multiplescaling}
  \centering
\resizebox{0.4\textwidth}{!}{\begin{tabular}{lcc}
\toprule
{\bf Inference-Time Strategy}   &{\bf Task SR (\%)}
\\ \toprule
Baseline & 23.81\\
Check-again  & 26.14 \\
Budget-forcing   &  24.81 \\
Best-of-$n$ & 25.03 \\
Check-again + Budget-forcing & 26.33\\
Check-again + Best-of-$n$ & 27.36\\
\bottomrule
\end{tabular}}
\end{table}

\subsection{\methodname{} and Online Filtered BC Hyperparameters for Preliminary Experiments}
\label{app:filteredbc}
We use the following hyperparameters to obtain the training curves in Figure~\ref{fig:traindynamics}.  During training, the \texttt{vision\_tower} of Gemma 3 is kept frozen because it is frozen during pretraining. Other hyperparameters can be found in our code.

\begin{itemize}
           \item num\_iteration: 10

    \item actor\_epochs: 1 \# number of epochs to update the actor
    \item rollout\_size: 512
    \item num\_update\_sample\_per\_iteration: 512
        \item lr: 1e-6

    \item optimizer: AdamW
    \item scheduler: WarmupCosineLR
        \item batch\_size: 4
    \item grad\_accum\_steps: 2  
    \item training\_gpu\_size: 4
    \item eval\_horizon: 30
\end{itemize}
For multiplicative curriculum, we use  the schedule: 10, 20, 30, 30, ... For additive curriculum, we use the schedule: 10, 11, 12, 13, ...

\subsection{\methodname{} Hyperparameters for Full WebArena Experiments}
\label{app:fullwebarena}
We use the following hyperparameters to obtain the full WebArena results for Table~\ref{table:webarenafull}. 

\begin{itemize}
           \item num\_iteration: 10
\item horizon\_schedule: 10, 20, 20, 30, 30, 30, ...
    \item actor\_epochs: 1 \# number of epochs to update the actor
    \item rollout\_size: 512
    \item num\_update\_sample\_per\_iteration: 512
        \item lr: 1e-6

    \item optimizer: AdamW
    \item scheduler: WarmupCosineLR
        \item batch\_size: 4
    \item grad\_accum\_steps: 2
    \item training\_gpu\_size: 4
    \item eval\_horizon: 30
\end{itemize}
\section{WebVoyager Experiments}

\subsection{Task Generator \& Evaluator Prompt}
\label{app:webvoyager:evalprompt}

\begin{tcolorbox}[
  colback=white,
  colframe=darkgray,
  title=Task Generator Prompt,
  fonttitle=\bfseries\centering,
  top=2mm,
  bottom=2mm,
  arc=3mm,
  breakable
]
You are a website exploration assistant tasked with discovering potential tasks on websites. These tasks should be similar to a user-specified task and aim to complete some high-level goals such as booking restaurants in a website. Your goal is to freely explore websites and propose tasks similar to a given set of examples. For each iteration, you'll receive:

- An observation with the webpage's accessibility tree

- A screenshot showing numerical labels in the TOP LEFT corner of web elements

You will then generate possible tasks while exploring the website. You should imagine tasks that are likely proposed by a most likely user of this website. You'll be given a set of examples for reference, but you must not output tasks that are the same as the given examples. The generated tasks must be realistic and at least require 3 steps to complete. It cannot be too simple.

\#\# Response Format and Available Actions

Your reply for each iteration must strictly follow this format:

Thought: Analyze the current webpage thoroughly to guide your exploration. Examine the webpage's structure, content, and interactive elements to identify potential tasks that users might perform on this site. Decide whether you want to keep exploring or output some tasks

Tasks: If you think you are ready to generate some tasks, output them in the following format (note that different tasks are separated with double semicolons): GENERATE [task1;answer1;;task2;answer2]

Action: Then, to continue with your exploration, choose ONE of the following action formats:

- Click [numerical\_label] - Click a specific element

- Type [numerical\_label] [content] - Input text into a field

- Scroll [up/down] - Navigate the page vertically

- GoBack - Return to previous webpage

Examples:

Click [8]

Type [22] [Boston]

Scroll [down]

GENERATE [Find the company's phone number;(555) 123-4567;;Locate the price of the basic subscription plan;\$19.99/month]

Your final output should look like:

Thought: ...

Tasks: GENERATE [...] (this is optional, only generate when you are confident)

Action: ...

\#\# Critical Guidelines

\#\#\# Action Rules

- Use either screenshot or accessibility tree to obtain the numerical\_label

- For text input, use Type action directly (no need to click first)

- Ensure proposed tasks are diverse and demonstrate different aspects of the website. The tasks must have diverse difficulty and require different number of steps (3-20) to complete.

- Tasks should be clear, specific, achievable, and self-contained. It cannot be too general, e.g., related to \"any post\", \"any product\", \"any place\". It must not depend on any context or actions that you have performed, i.e., you must assume zero prior knowledge when someone wants to complete the task

- Your task should be objective and unambiguous. The carry-out of the task should NOT BE DEPENDENT on the user's personal information such as the CURRENT TIME OR LOCATION

- Your tasks should be able to be evaluated OBJECTIVELY. That is, by looking at the last three screenshots and the answer provided by an agent, it should be possible to tell without ambiguity whether the task was completed successfully or not

- Answers should be precise (e.g., exact prices, specific information, exact text)

- Your should output both operational tasks (the goal is to complete some steps) and information retrieval tasks (the goal is to find some answer to return)

- You must refer to the examples given and mimic the complexity and task structure. See how these tasks are self-contained and realistic

- Your proposed task cannot be a single action like click, type! Tasks like 'Determine the number of uses for that term' is unacceptable because it is ambiguous as a stand-alone task; 'Uncheck Use system value' is unacceptable because it is not a complete task; 'Locate the total revenue for the last month' is unacceptable because 'last month' is ambiguous;

After each action, you'll receive a new observation. Continue exploring and generating tasks.

Here're some examples:
\{example\}

Current URL: \{url\}

Screenshot of current viewpoint: attached

Accessibility tree of current viewpoint:
\{accessibility\_tree\}
\end{tcolorbox}

\begin{tcolorbox}[
  colback=white,
  colframe=darkgray,
  title=Evaluator Prompt,
  fonttitle=\bfseries\centering,
  top=2mm,
  bottom=2mm,
  arc=3mm,
  breakable
]

You are an expert in evaluating the performance of a web navigation agent. The agent is designed to help a human user navigate a website to complete a task. Your goal is to decide whether the agent's execution is successful or not.

As an evaluator, you will be presented with three primary components to assist you in your role:

1. Web Task Instruction: This is a clear and specific directive provided in natural language, detailing the online activity to be carried out.

2. Result Response: This is a textual response obtained after the execution of the web task. It serves as textual result in response to the instruction.

3. Result Screenshots: This is a visual representation of the screen showing the result or intermediate state of performing a web task. It serves as visual proof of the actions taken in response to the instruction.

-- You SHOULD NOT make assumptions based on information not presented in the screenshot when comparing it to the instructions.

-- Your primary responsibility is to conduct a thorough assessment of the web task instruction against the outcome depicted in the screenshot and in the response, evaluating whether the actions taken align with the given instructions.

-- NOTE that the instruction may involve more than one task, for example, locating the garage and summarizing the review. Failing to complete either task, such as not providing a summary, should be considered unsuccessful.

-- NOTE that the screenshot is authentic, but the response provided by LLM is generated at the end of web browsing, and there may be discrepancies between the text and the screenshots.

-- Note that if the content in the Result response is not mentioned on or different from the screenshot, mark it as not success.

-- NOTE that the task may be impossible to complete, in which case the agent should indicate this in the response. CAREFULLY VERIFY THE SCREENSHOT TO DETERMINE IF THE TASK IS IMPOSSIBLE TO COMPLETE. Be aware that the agent may fail because of its incorrect actions, please do not mark it as impossible if the agent fails because of its incorrect actions.

You should explicit consider the following criterion:

- Whether the claims in the response can be verified by the screenshot. E.g. if the response claims the distance between two places, the screenshot should show the direction. YOU SHOULD EXPECT THAT THERE IS A HIGH CHANCE THAT THE AGENT WILL MAKE UP AN ANSWER NOT VERIFIED BY THE SCREENSHOT.

- Whether the agent completes EXACTLY what the task asks for. E.g. if the task asks to find a specific place, the agent should not find a similar place.

In your responses:

You should first provide thoughts EXPLICITLY VERIFY ALL THREE CRITERION and then provide a definitive verdict on whether the task has been successfully accomplished, either as 'SUCCESS' or 'NOT SUCCESS'.

A task is 'SUCCESS' only when all of the criteria are met. If any of the criteria are not met, the task should be considered 'NOT SUCCESS'.
\end{tcolorbox}

\subsection{Agent Prompt}

\begin{tcolorbox}[
  colback=white,
  colframe=darkgray,
  title=WebVoayager,
  fonttitle=\bfseries\centering,
  top=2mm,
  bottom=2mm,
  arc=3mm,
  breakable
]
Imagine you are a robot browsing the web, just like humans. Now you need to complete a task. In each iteration, you will receive an observation that includes the accessibility tree of the webpage and a screenshot of the current viewpoint. The accessbility tree contains information about the web elements and their properties. The screenshot will feature numerical labels placed in the TOP LEFT corner of web elements in the current viewpoint.
Carefully analyze the webpage information to identify the numerical label corresponding to the web element that requires interaction, then follow the guidelines and choose one of the following actions:

1. Click a web element.

2. Delete existing content in a textbox and then type content. 

3. Scroll up or down the whole window.

4. Go back, returning to the previous webpage.

5. Navigate to Bing's homepage.

6. Answer. This action should only be chosen when all questions in the task have been solved.

Correspondingly, action should STRICTLY follow the format specified by one of the following lines:

Click [numerical\_label]

Type [numerical\_label] [content]

Scroll [up/down]

GoBack

Bing

ANSWER [content]

Some examples are:

Click [8]

Type [22] [Boston]

Scroll [down]

Bing

ANSWER [06516]

Key guidelines you MUST follow:

* Action guidelines *

1. The predicted action should be based on elements as long as it's accessibility tree OR screenshot. Sometimes, accessibility tree or screenshot captures more elements than the other, but it's fine to use either one.

2. To input text for search bars, no need to click textbox first, directly type content. After typing, the system automatically hits 'ENTER' key.

3. When a complex task involves multiple questions or steps, select 'ANSWER' only at the very end, after addressing all of these questions or steps. Double check the formatting requirements in the task when ANSWER. Always think twice before using 'ANSWER' action!!!

4. When specifying the content for 'Type' and 'ANSWER' actions, be sure to wrap the content with '[]'.

5. Use `GoBack` to return to the previous state, use it when you find the previous action incorrect. 

6. When you see a pop-up page, you should immediately `GoBack` to the previous page.

7. Use `Bing` when you need to navigate to a different website or search for new information.

Your reply should strictly follow the format:

Thought: Your reasoning trace. A good practice is to follow this format:

- Observation summary: where are you at now? list all elements that are related to the task goal. e.g. if you're trying to filter something out, list all filters visible.

- Planning: what sequence of actions do you need take to achieve the task goal? give a high-level overview of the steps you need to take.

- Possible actions: to achieve that plan, what are potential actions you need to do immediately and what's their effect? List at least 3 actions and analyze each of them.

Action: Based on this reasoning, identify the single most optimal action. You should output it in the format specified above ("...STRICTLY follow the format...").

After you issue an action, the user will execute it and provide a new observation. Now solve the following task.

Task: \{task\_goal\}

Current URL: \{url\}

Screenshot of current viewpoint: attached

Accessibility tree of current viewpoint:
\{accessibility\_tree\}
\end{tcolorbox}

\subsection{Experiment Details}
\label{app:webvoyager:hyperparam}
We use the following hyperparameters to obtain the WebVoyager results. 

\begin{itemize}
    \item num\_iteration: 12
    \item horizon\_schedule: 10, 20, 20, 30, 30, 30, 30, 30, 30, 30, 30, 30, 30
    \item actor\_epochs: 1 \# number of epochs to update the actor
    \item rollout\_size: 512
    \item num\_update\_sample\_per\_iteration: 512
    \item lr: 4e-6

    \item optimizer: AdamW
    \item scheduler: WarmupCosineLR
        \item batch\_size\_per\_gpu: 4
    \item grad\_accum\_steps: 2   
    \item training\_gpu\_size: 6
    \item eval\_horizon: 30 \# note that train horizon is different for different methods, but evaluation horizon is kept the same
\end{itemize}

In~\Cref{fig:webvoyager:train}, the green area starts from iteration 6, because this is the first iteration where it's possible to have all trajectories sampled for training have a horizon of $H=30$.

\newpage
\subsection{Case Studies: Strengths 1}
\label{app:case:strength1}

Task: Locate a recipe for an American apple pie on Allrecipes with a rating of at least 4 stars and more than 50 reviews. Note the maximum temperature mentioned in the Directions.

Fully trained agent explores: 
\begin{figure}[H]
    \centering
    \includegraphics[width=0.8\linewidth]{figs/687-complete.pdf}
\end{figure}

\newpage
Early-stage agent prefers exploitation:
\begin{figure}[H]
    \centering
    \includegraphics[width=0.45\linewidth]{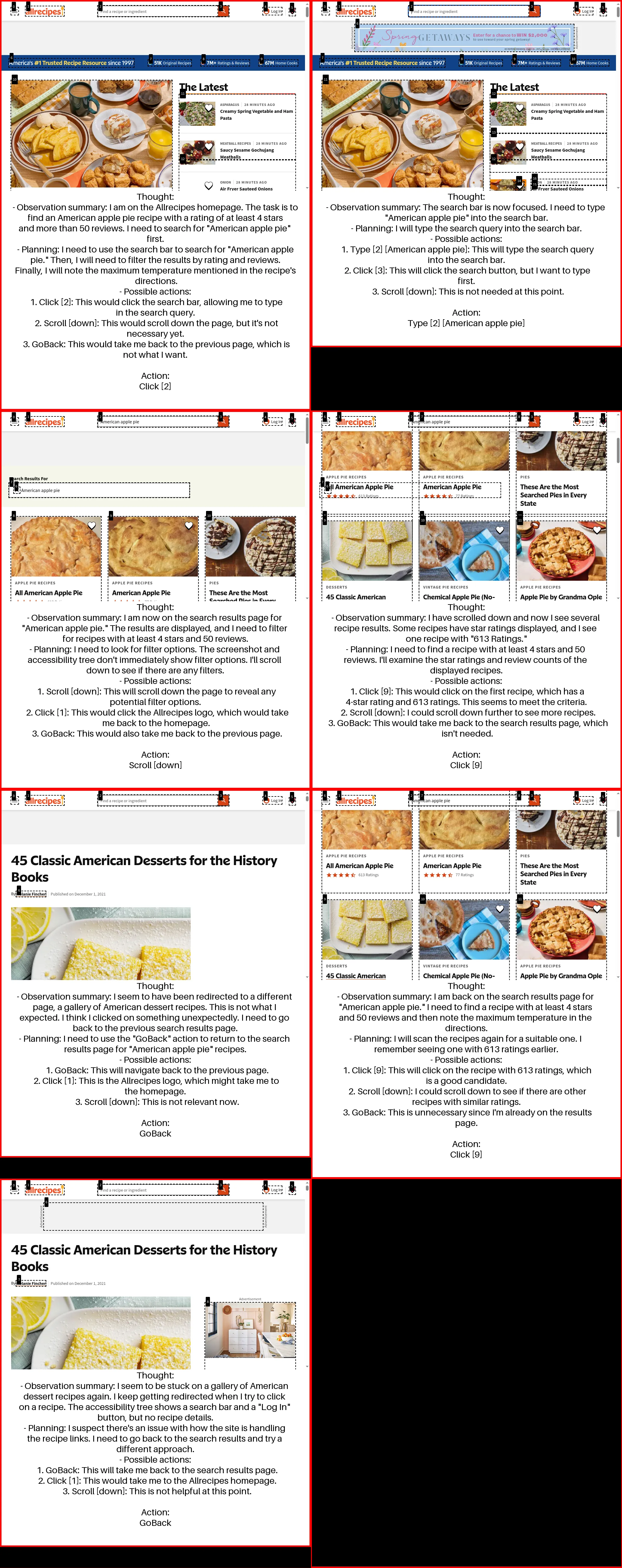}
\end{figure}

\newpage

\subsection{Case Studies: Strengths 2}
\label{app:case:strength2}

Task: Identify the latest top-trending open-source project in the category of `Machine Learning' on GitHub, and check the number of stars it has received.

\begin{figure}[H]
    \centering
    \includegraphics[width=0.35\linewidth]{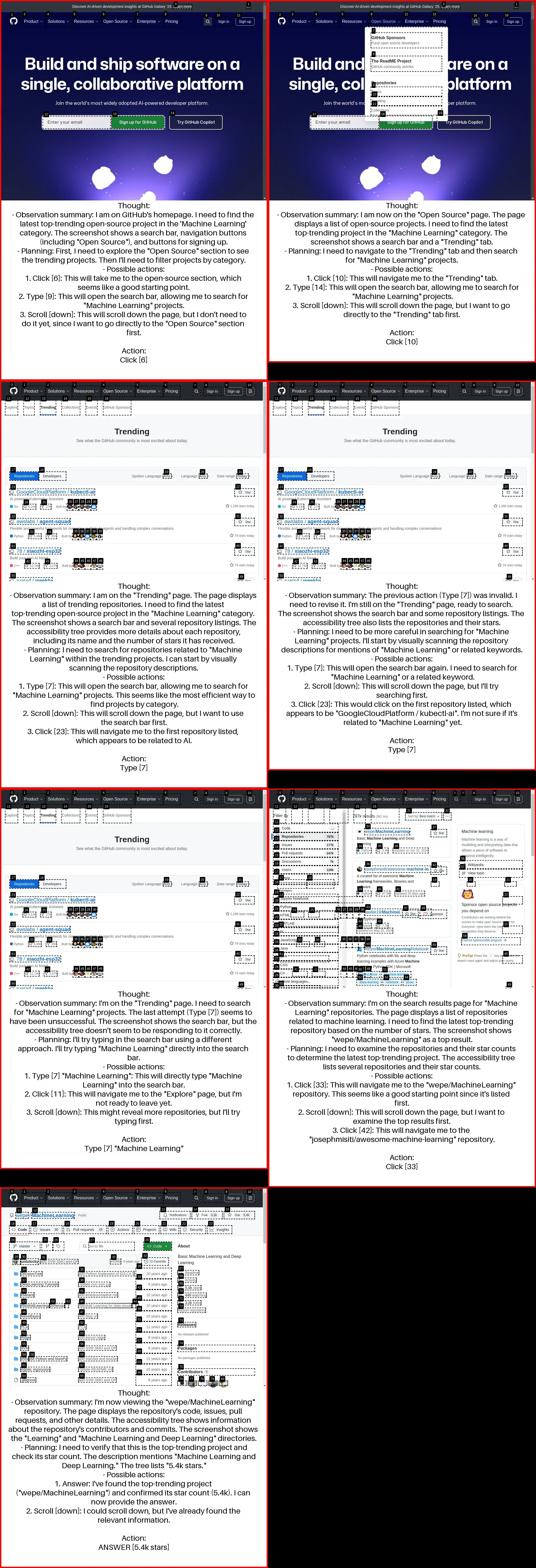}
\end{figure}

\newpage
\subsection{Case Studies: Fail Modes 1}
\label{app:case:fail1}

Task: On Apple's website, how many different types of keyboards are available when customizing your 14-inch MacBook Pro?
\begin{figure}[H]
    \centering
    \includegraphics[width=0.80\linewidth]{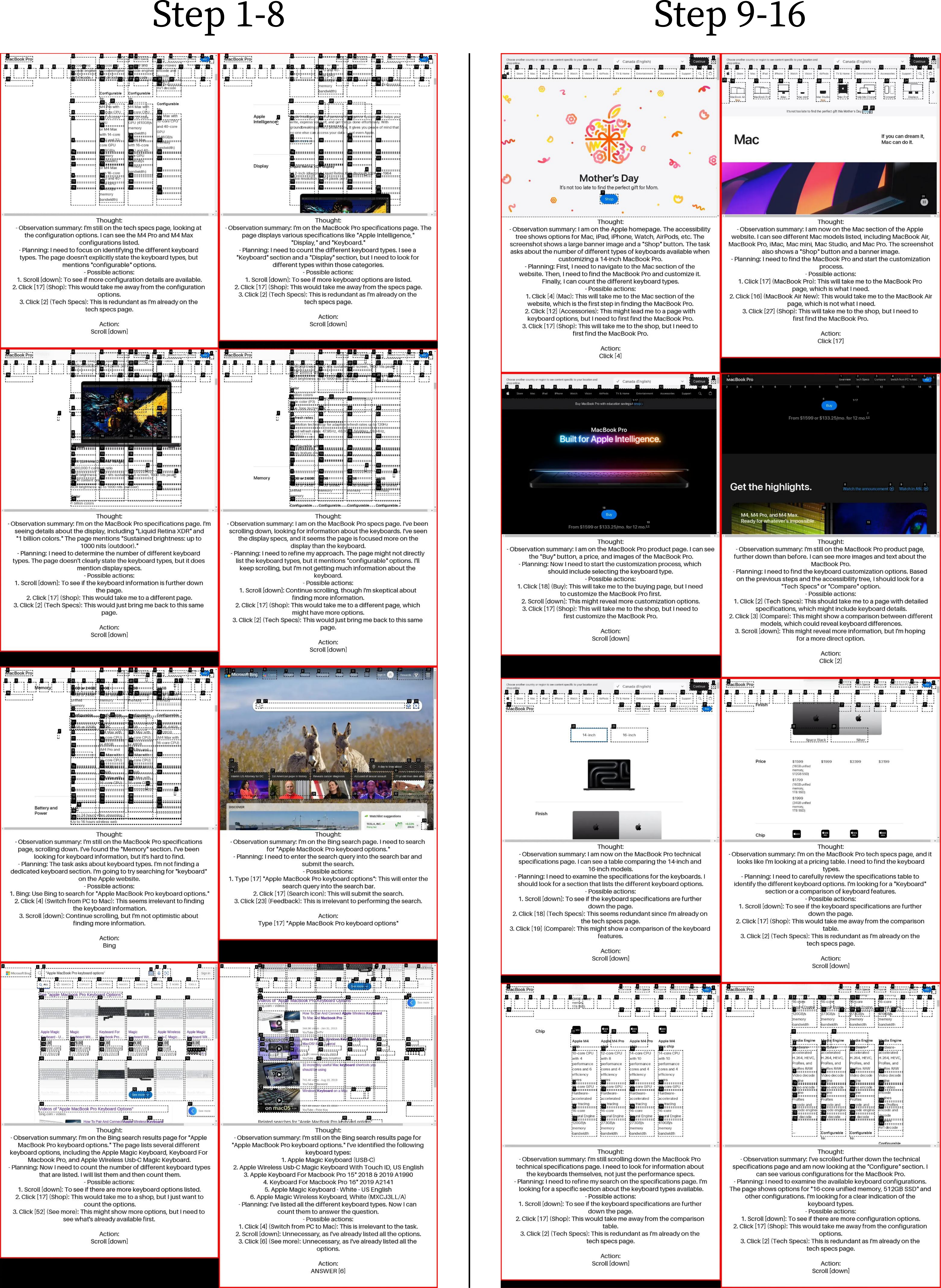}
\end{figure}

\newpage

Task: Which university maintains and manages ArXiv. Accessing the university's website from ArXiv, how many undergraduate students are currently at the university.

\begin{figure}[H]
    \centering
    \includegraphics[width=0.5\linewidth]{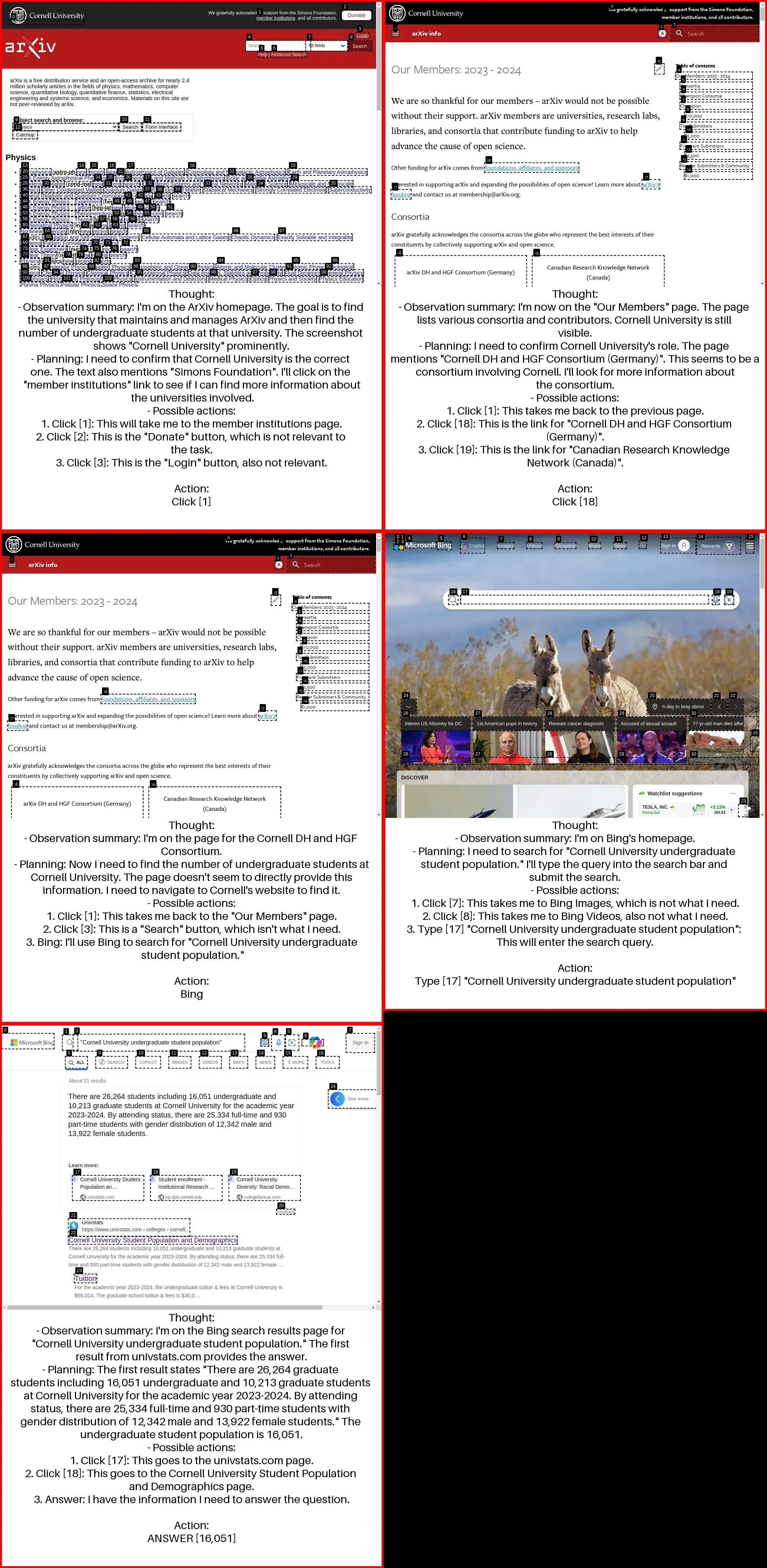}
\end{figure}

\newpage
\subsection{Case Studies: Fail Modes 2}
\label{app:case:fail2}
Task: Identify a new open-source project on GitHub related to `AI agriculture' that created in 2022, and note its main programming language and description.

\begin{figure}[H]
    \centering
    \includegraphics[width=0.4\linewidth]{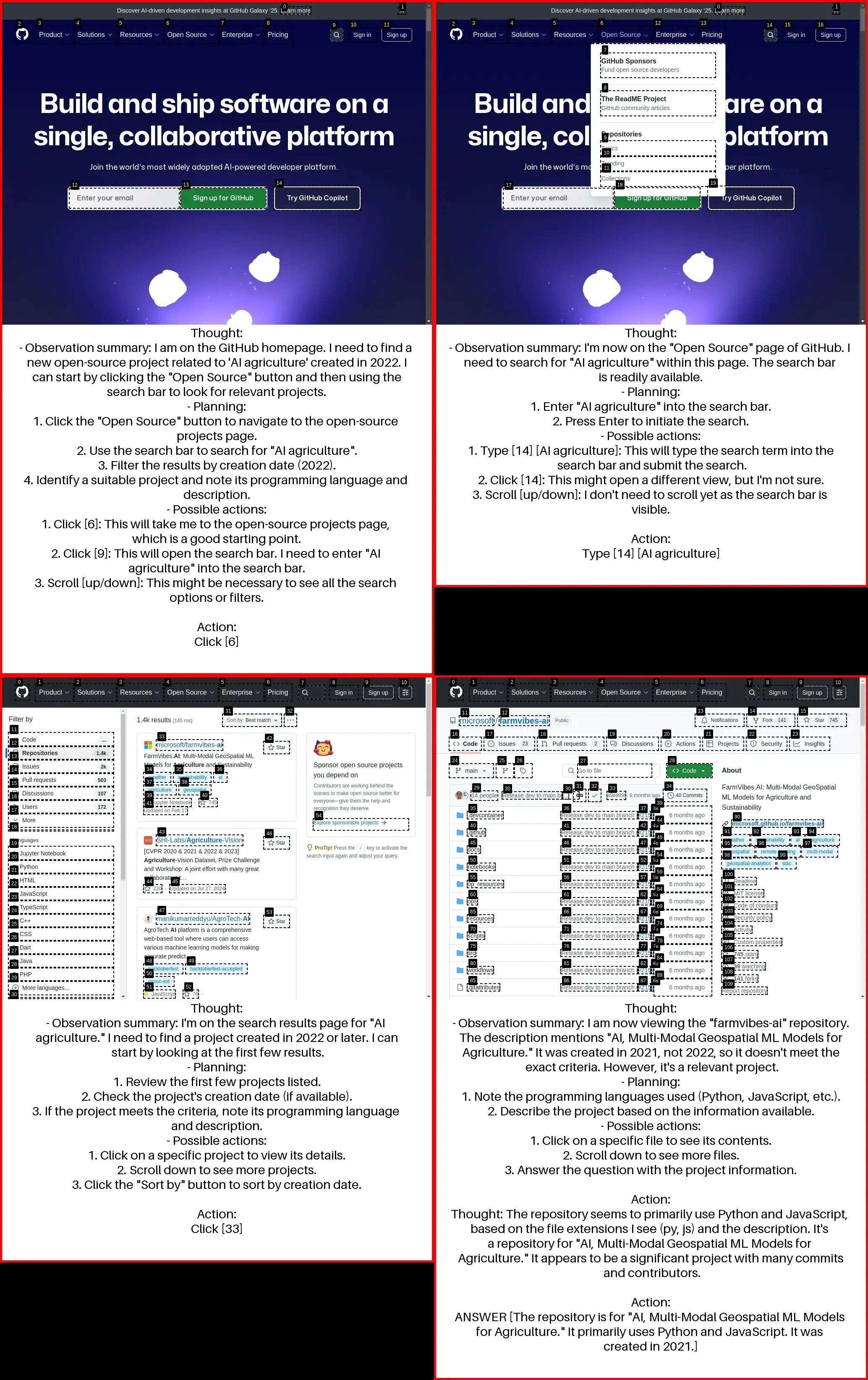}
\end{figure}

\end{document}